\documentclass[11pt]{article}

\usepackage{fullpage}

\usepackage{booktabs} 

\usepackage{graphicx}
\usepackage[tight]{subfigure}

\usepackage{amsmath}
\usepackage{tcolorbox}

\usepackage{amsthm,tcolorbox}

\newtheoremstyle{stmt}
 {3pt}
 {3pt}
 {\itshape}
 {}
 {\bfseries}
 {}
 {1em}
 {(\thmnumber{#2})}

\theoremstyle{stmt}
\newtheorem{stmt}{}[section]
\newtheorem{stmt*}[stmt]{}

\tcolorboxenvironment{stmt*}{
  frame empty,
  colback=black!15!white,
  grow to left by=6pt,
  grow to right by=6pt,
  left=1.6pt,
  right=1.6pt,
  arc=0pt
}

\begin{document}

\renewcommand{\proof}[1]{
{\noindent {\it Proof.} {#1} \rule{2mm}{2mm} \vskip \belowdisplayskip}
}

\newcommand{\prevs}[2]{
{\vskip 0.1in \noindent {\em Proof of \rf{#1}.} {#2} \rule{2mm}{2mm}
\vskip \belowdisplayskip}
}

\def\eps{{\varepsilon}}

\def\R{{\bf R}}

\renewcommand\^[1]{^{\langle#1\rangle}}
\renewcommand\-[1]{^{(#1)}}

\newcommand{\xhdr}[1]{\paragraph{\bf #1.}}

\newcommand{\omt}[1]{}

\newcommand{\rf}[1]{(\ref{#1})}


\def\Prf{{\rm Pr}}
\def\ev{{\cal E}}
\def\evf{{\cal F}}

\newcommand{\Prb}[1]{
\Prf\left[{#1}\right]
}

\newcommand{\Prg}[2]{
\Prf\left[{#1}~|~{#2}\right]
}

\newcommand{\Exp}[1]{
E\left[{#1}\right]
}

\newcommand{\Expg}[2]{
E\left[{#1}~|~{#2}\right]
}

\newcommand{\Prbt}[1]{
\Prf\left[\mbox{\em #1}\right]
}

\newcommand{\Prgt}[2]{
\Prf\left[\mbox{\em #1}~|~\mbox{\em #2}\right]
}


\newcommand{\avgf}[2]{
 \overline{{#1}|{#2}}
}

\def\gp{\gamma}
\def\x{{x}}
\def\xb{\overline{\x}}
\def\sb{\sigma}
\def\indgp{{\bf 1}_D}
\def\dimbd{B}
\def\gfn{\theta}
\def\ms{\mu}
\def\rv{Y}
\def\adm{A}
\def\fval{y}
\def\basepsi{\psi_0}
\def\gpdep{\chi}

\def\efd{v}
\def\eqd{w}
\def\efc{V}
\def\eqc{W}

\def\efn{\efc^*}
\def\eqn{\eqc^*}

\def\v{{v}}
\def\cv{y}
\def\yset{M}
\def\maxg{g_f^*}
\def\btau{\overline{\tau}}

\def\fa{\ms_A}
\def\fd{\ms_D}



\def\ma{\alpha}
\def\md{\delta}

\def\rvd{P}
\def\rva{Q}
\def\rvval{u}

\def\rvdprob{p}
\def\rvaprob{q}
\def\cdd{\rvdprob^+}
\def\cda{\rvaprob^+}




\title{Simplicity Creates Inequity: \\ Implications for Fairness, Stereotypes, and Interpretability}

\author{
Jon Kleinberg\thanks{Departments of Computer Science and Information Science, Cornell University} \and 
Sendhil Mullainathan\thanks{University of Chicago Booth School of Business}
}

\date{}
\maketitle

\begin{abstract}
Algorithms are increasingly used to aid, or in some cases supplant,
human decision-making, particularly for decisions that hinge on
predictions. As a result, two additional features in addition to
prediction quality have generated interest: (i) to facilitate human
interaction and understanding with these algorithms, we desire
prediction functions that are in some fashion simple or interpretable;
and (ii) because they influence consequential decisions, we also want
them to produce equitable allocations.
We develop a formal model to explore the relationship between
the demands of simplicity and equity. Although the two concepts appear to be
motivated by qualitatively distinct goals, we show a
fundamental inconsistency between them. Specifically, we formalize
a general framework for producing simple prediction functions,
and in this framework we establish two basic results.
First, every simple
prediction function is strictly improvable: there exists a more
complex prediction function that is both strictly more efficient and
also strictly more equitable. Put another way, using a simple
prediction function both reduces utility
for disadvantaged groups and reduces overall welfare
relative to other options.
Second, we show that simple prediction functions
necessarily create incentives to use information about
individuals' membership in a disadvantaged group --- incentives
that weren't present before simplification, and that
work against these individuals.
Thus, simplicity transforms disadvantage into 
bias against the disadvantaged group.
Our results are not only about algorithms but about any
process that produces simple models, and as such they connect to the
psychology of stereotypes and to an earlier economics literature on
statistical discrimination.
\end{abstract}


\section{Introduction}
\label{sec:intro}

Algorithms can be a powerful aid to decision-making --- particularly
when decisions rely, even implicitly, on predictions
\cite{kleinberg2015prediction}. We are already seeing algorithms play
this role in domains including hiring, education, lending, medicine,
and criminal justice
\cite{chouldechova-child-hotline,kleinberg-bail-qje,rockoff-teacher-recruiting,stevenson-assessing-risk-assessment}.
Across these diverse contexts, the role for algorithms follows a similar template: 
{\em applicants} present themselves to be evaluated by a
{\em decision-maker} who chooses an accept/reject outcome for each applicant ---
for example, whether they are hired, admitted to a selective school,
offered a loan, or released on bail.
The final decision-making authority in these situations typically 
rests with a human being or a committee of human beings. But because the decisions turn on a prediction of some underlying quantity (such as crime risk in the case of bail or default risk in the case of a loan), decision-makers are beginning to rely on the assistance of 
algorithms that map features of each applicant to a numerical prediction.

As is typical in machine learning applications,  \emph{accuracy}, evaluated by some measure of admitted applicants' future performance, is an important measure. In these high-stakes policy contexts, though, two additional considerations prove important as well, as highlighted by recent work:
\begin{itemize}
\item {\em Fairness and equity.}  Certain groups
in society are {\em disadvantaged} in clear-cut quantitative ways ---
on average they graduate from less-resourced educational institutions,
live in areas with reduced economic opportunities, and face other
socioeconomic challenges in aggregate. 
Will an algorithmic approach, based on these underlying measures, 
perpetuate (or even magnify) the underlying disadvantage? Or could we use algorithms to increase equity between groups?
\cite{barocas-big-data-disparate,corbett-davies-critical-review-fair,dwork-fairness-awareness,feldman-certifying-disparate}
\item {\em Interpretability.}  Algorithms tend to  result in complex models that are hard for human beings
to comprehend.  Yet in these domains, humans work intimately with them.  Can a decision-maker derive 
understanding from such an algorithm's output, or are they forced
to treat it as a ``black box'' that produces pure numerical predictions
with no accompanying insight?  And similarly, can an applicant derive
any understanding about the basis for the algorithm's prediction
in their particular case?
\cite{doshi-velez-science-interp,lipton-mythos-interp,zeng-ustun-interpretable}
\end{itemize}

Fairness and interpretability are clearly distinct issues, 
but it is natural to suspect that there may be certain interactions
between them.
A common theme in the literature on interpretability is the
possibility that interpretable models can be more easily examined
and audited for evidence of unfairness or bias; as
one of many examples of this theme,
Doshi-Velez and Kim argue that
``interpretability can assist in qualitatively ascertaining whether
other desiderata --- such as fairness, privacy, reliability, robustness,
causality, usability and trust --- are met''
\cite{doshi-velez-science-interp}.
Nor is this point purely an issue in academic research:
the formulation of the
European Union General Data Protection Regulation (GDPR)
reinforces earlier EU regulations asserting that individuals have
a ``right to explanation'' when they are affected 
by algorithmic decision-making. 
The technical implications of these guidelines are not yet fully clear,
but their premise situates interpretable decisions
as a component of fair outcomes
\cite{goodman-gdpr,lipton-mythos-interp}.



\xhdr{The present work: A basic tension between fairness and simplicity}
There are many ways in which interpretability may be able to help promote
fairness --- they might be more easily analyzable and auditable, as
noted above; and the activity of constructing an interpretable rule,
depending how it is carried out, 
may be able to engage more participants in the process.

But there has been arguably less exploration of what, if anything,
we give up with respect to fairness and equity
when we pursue interpretable rules.
Here we consider a set of questions in this direction,
focusing in particular on the role of {\em simplicity} in 
the construction of prediction rules.
Simplification is one of the common strategies employed in the
construction of interpretable models, and this is for natural reasons.
A primary source of model complexity is ``width'' --- the number of
variables, or applicant features, that are used. Humans typically
struggle to understand very wide models, and so to create
interpretable or explainable models, many standard approaches seek, at some
level, to reduce the number of variables that go into in any one decision.
There are many ways to do this: for example,
we could project the space of features
onto a small number of the most informative variables;
we could enumerate short ``rules'' that only depend
on a few variables;
we could construct a shallow 
decision tree that only consults a small number of variables 
on any path from its root to a leaf.
Despite the diversity in these approaches, they follow
a common principle: they all {\em simplify}
the underlying model by combining distinguishable 
applicants into larger sets and making a common decision at 
the level of each set.

Our main results show that 
when one group of individuals is disadvantaged with respect to another,
there is a precise sense in which 
the process of simplifying a model will necessarily 
hurt natural measures of fairness and equity.

\xhdr{The present work: Summary of results}
The exact statement of our results will be made precise via
the model that we develop starting in the next section,
but roughly speaking they proceed as follows.
We first formalize the above notion of simplicity:
given applicants with feature vectors, and a function for ranking
applicants in terms of their feature vectors,
we say that a {\em simplification} of this function is
a partitioning of the feature vectors into {\em cells},
such that each cell is obtained by fixing the values of
certain dimensions in the feature vector and leaving others unrestricted.
A fixed value is assigned to each cell, computed as the 
average value of all the applicants who are mapped to the cell.
This generalizes, for example, the structure we obtain when we
project the function onto a reduced number of variables, or
group applicants using a short decision tree or decision list.
We say that a simplification is {\em non-trivial} if at least
some of its cells average together applicants of different underlying values.

We show that under these definitions, when one group
experiences disadvantage relative to another, 
non-trivial simplifications of a function exhibit two problems:
they can be strictly improved, and they create incentives to
explicitly use information about an applicant's membership in
a disadvantaged group.
We describe each of these problems in more detail.

First, we will prove that any non-trivial 
simplification of a function is {\em strictly improvable}:
it can be replaced with a more complex function which produces an outcome
that is simultaneously more accurate overall and also more equitable 
toward the disadvantaged group.
Thus, whatever one's preferences for accuracy and equity in a model,
the complex function dominates the simple one.
In the language of optimization, this means that every simple
model is {\em strictly Pareto-dominated} --- since it can be improved in both
accuracy and equity simultaneously, it never represents the best
trade-off between these two criteria.

Now, it is intuitively natural that one can improve on the  {\em accuracy} of a simple model: much of the literature in this area is organized in terms of a trade-off between interpretability and performance.
But as the above discussion illustrates, it has generally been imagined that we are agreeing to this trade-off because interpretability brings collateral benefits like the promotion of fair and equitable outcomes.
This is the aspect of the earlier arguments 
that our result calls into question:
in a formal sense, {\em any} attempt at simplification in fact creates inequities that a more complex model could eliminate while also improving performance.
Achieving interpretabilty through simplification sacrifices not only performance but also equity.

Simplifying a function also introduces a second problem.
Suppose that the true function for ranking applicants does not depend on
group membership --- applicants who differ only in their group
receive identical evaluations by this true function.
As a result, the ranking by the true (complex) function 
would be the same whether or not group membership was known.
We show, however, that simple functions that do not use group membership
can always be made more accurate if they are given access to 
group membership information.
Moreover, this improvement in accuracy comes at the cost of reducing
equity toward the disadvantaged group: faced with two otherwise
identical applicants, the one from the disadvantaged group
would be ranked lower.
This creates a troubling
contrast: with the original function, a decision-maker concerned with
maximizing accuracy had no interest in which group an applicant
belonged to; but once we simplify the function in any non-trivial way,
the decision-maker suddenly has an interest in using group membership
in their ranking, and in a way that hurts the disadvantaged group. Put
informally, simple functions create an incentive to
``seek out'' group membership for
purposes of discriminating against the disadvantaged group,
in a way that more complex functions don't.
Simplification transforms disadvantage into explicit bias.

A concrete example helps illustrate 
these two ways in which simplicity sacrifices equity.
Suppose that a college, to simplify its
ranking of applicants, foregoes the use of admissions essays for all students.
(This is in keeping with the type of simplification discussed above:
in the representation of each applicant, the college is grouping
applicants into cells by
projecting out the dimension corresponding to the quality of the essay.)
In doing so, it harms those disadvantaged students with excellent
essays: they now have no way of showing their skill on this dimension.
Moreover, suppose that the disadvantaged group has a lower fraction of
applicants with strong essays, precisely because students from the
disadvantaged group come from less-resourced educational institutions.
Then the college's simplified evaluation creates a perverse incentive 
to use an applicant's advantaged/disadvantaged status 
as an explicit part of the ranking, because group membership
conveys indirect information about the average quality of the (unseen) essay.
Simplification not only means that members of the
disadvantaged group have fewer ways
of showing their skill; it also transforms
group status into a negative feature.
Though this is one specific example, the machinery of our proof shows 
that such problems are endemic to all forms of simplification
in this style.

Recent empirical work also provides revealing analogues to these
results. In particular, recent studies have investigated policies that
seek to limit employers' access to certain kinds of job applicant
information. For example, ``ban-the-box'' policies prevent employers
from asking whether applicants have criminal records, with the goal of
helping those applicants with prior criminal convictions. A striking
study of Agan and Starr argued that such policies can have unintended
consequences: through a large field experiment measuring callbacks for
hiring they found that when localities implemented ban-the-box
policies, racial disparities increased significantly
\cite{agan-ban-the-box}. One of the main interpretations of this
finding can be understood in terms of simplification: by eliminating a
feature of an applicant (existence of a criminal record) that is
correlated with membership in a disadvantaged group and would have
been used in the decision, the policy creates an unintended incentive
to make greater explicit use of membership in this disadvantaged group
instead as part of the decision.

\xhdr{Scope of Results}
The scope of our results are both more specific and more general than they
may initially appear.  First, it is important to keep in mind that our
work is based on a particular definition of simplicity.  While the
definition is quite general, and captures many of the formalisms
in wide use, there are other ways in which simplicity could be formulated,
and these alternative formulations may lead to different
structures than what we describe here. 
And beyond this, the notion of interpretability is more general still;
simplification is only one of the standard strategies used in 
developing interpretable models.
Thus, in addition to the
results themselves, we hope our work can help lay the groundwork for
thinking about the interaction of fairness, simplicity, and interpretability
more generally.

At the same time, our notion of simplicity can be motivated in 
several independent ways beyond the initial considerations of interpretability.
In particular, we may be using a simple model because
more complicated models are computationally
complex and time-consuming to fit. Data collection costs could lead to
measuring fewer variables. At an even more fundamental level, machine learning
methods naturally give rise to simplicity.  To address over-fitting,
procedures that estimate high-dimensional models typically choose a simpler
model that fits worse in-sample but performs better out-of-sample
\cite{hastie-stat-learning-book}.
For example, the process of growing a decision tree generally
has a stopping condition that prevents the number of instances 
being mapped to a single node from getting too small;
further refinement of the tree may provide signal but will not be undertaken because the magnitude of that signal does not exceed 
a specified regularization penalty.  All these diverse motivations may provide important reasons to pursue simplification in a range of contexts.  
The central point of our framework here, however, is to identify a further cost inherent in these choices --- that simplification gives up some amount of equity.

Additionally, concerns about accuracy, fairness, and simplicity are relevant not just to algorithmic decisions but to purely human ones as well; and therefore much of our analysis  implies fundamental constraints on {\em any} system for decision-making, whether algorithmic or human. To the extent that human beings think in categories, these categories can be viewed as coarse simplifications and our results also apply to them \cite{mullainathan-categories,mullainathan-coarse-thinking}.
Indeed, our findings suggest a connection to an important issue
in human psychology --- the construction of {\em stereotypes}
\cite{greenwald-implicit-social-cognition,leyens-stereotypes-book}.
If we think of stereotyping as a process of taking distinguishable 
individuals and grouping them together so as to treat them similarly,
then our results show that when there is disadvantage between groups,
all ways of stereotyping will increase inequity.
And in this way, we also arrive at a possible counterweight to the
earlier intuitions about interpretability and its potential benefits
for fairness: 
requiring an algorithm to work with a model of reduced
complexity is effectively asking it to construct stereotypes from
the data, and this activity reduces not only performance but also equity.

\section{An Informal Overview of the Model}
\label{sec:informal}
It will be useful to provide an informal overview of the model before specifying it in detail. 

The model represents the process of {\em admissions} or {\em screening}:
we have a set of {\em applicants}, and we would like to admit a fraction $r$ of them, a quantity we will refer to as the {\em admission rate}. 
We can think of this process for example as one
of hiring, or lending, or admission to a selective school.
Each applicant is described by a {\em feature vector}, and 
they also belong to one of two {\em groups}: 
an {\em advantaged} group $A$ or a {\em disadvantaged} group $D$.
There is a function $f$ that maps each individual to their 
qualifications for purposes of admission: 
$f$ represents whatever criterion
we care about for the admission process.
We assume $f$ is an arbitrary function of an applicant's feature vector but does not depend on
group membership; if two applicants have the same feature vector
but belong to different groups, they have the same  $f$-value.
Thus, group membership has no true effect on someone's qualifications
as an applicant.
However, group $D$ does experience {\em disadvantage}, in the sense that a smaller proportion of the applicants in group $D$ have feature vectors
that produce large values of $f$.

Now, the basic task is to assign each applicant a {\em score}, so that applicants can be ranked in decreasing order of this score, and then the top $r$ fraction can be admitted.  The basic two measures that we would like to optimize in admissions
are the {\em efficiency} of the admitted set, defined as the average
$f$-value of the admitted applicants, and the {\em equity} of
the admitted set, defined as the fraction of admitted applicants who
come from group $D$.
Perhaps the most natural score to assign each applicant is their
true $f$-value; this makes sense from the perspective of efficiency, since
we are admitting a subset of the applicants whose $f$-values are as
high as possible.

But this is where the issue of simplification comes in.
It may be that $f$ is too complicated or impractical to work with,
or even to represent;
or we would like a more interpretable score; 
or perhaps we are hoping that by simplifying $f$ we might improve equity
(even at the cost of potentially reducing efficiency).
Thus, we consider simpler scores $g$, obtained by grouping sets of
feature vectors into larger {\em cells} of applicants who will be
treated the same, and assigning a 
group average score to all the applicants in a single cell.
Not every way of partitioning feature vectors into cells should be
viewed as ``simple''; some partitions, for example, would arguably
be more complicated to express than $f$ itself.
We thus think of $g$ as a {\em simple} score if each of its cells
has the following structured representation: we fix the values of
certain dimensions of the feature vector, taking all applicants
who match the values in these dimensions, and leaving the values of
all other dimensions unspecified.
As we discuss further in the next section,
many of the most natural formalisms for creating
simple or interpretable functions have this structure, and thus we
are abstracting a practice that is standard across multiple different methods.

\xhdr{Main Results}
From here we can informally state our first main result as follows: 
for every admission rule based on a simple function $g$, there
is a different function $h$ (possibly not simple), such that if
we admit applicants using $h$ instead of $g$, then both the resulting efficiency
and the resulting equity are at least as good for every admission rate $r$; 
and both are strictly better for some admission rate $r'$.
Thus, however our preferences for efficiency and equity are weighted,
we should prefer $h$ to $g$.
In other words, simple rules are {\em strictly Pareto-dominated}: 
a simple rule never represents the best trade-off
between efficiency and equity, since it can be simultaneously improved
in both respects.

The proof of this first result, informally speaking, starts with an
arbitrary simple function and looks for a cell that either contains
applicants from group $D$ whose $f$-values are above the average score
in the cell, or contains 
applicants from group $A$ whose $f$-values are below the average score
in the cell.  
In either case, by separating these applicants out into a distinct cell,
we end up with a new function that has moved forward in its ranking
applicants with higher average $f$-values and higher representation 
from group $D$, resulting in a strict improvement.
The key issue in the proof is to show that one of these improving operations
is always possible, for any simple function.

We also show a second result, using the notion of a 
{\em group-agnostic} simplification of $f$ --- a score $g$ based on
combining feature vectors into cells in such a way that 
applicants who differ only in group membership are mapped to the same cell.
We show that when we incorporate knowledge of group membership into $g$ ---
by ``splitting'' each cell into distinct cells for the applicants from
groups $A$ and $D$ respectively --- the efficiency of the resulting
admission rule goes up, and the equity goes down.
We conclude that even though group membership is irrelevant to the true
value of $f$, any group-agnostic simplification of $f$ creates an incentive
for a decision-maker to use knowledge of group membership --- an
incentive that wasn't present before simplification, and one
that hurts the disadvantaged group $D$.

With this as the overview, we now give a more formal description of the model.

\section{A Model of Simplicity and Equity}
\label{sec:model}

\subsection{Feature Vectors, Productivity, and Disadvantage}

We begin with feature vectors.
Each applicant is described by a feature vector consisting
of $k$ {\em attributes} $\x\^1, \ldots, \x\^k$, where each $\x\^i$ 
is a Boolean variable taking the value $0$ or $1$.
(Later we will see that the assumption that the variables are 
Boolean is not crucial, but for now it is useful for concreteness.)
As discussed above,
each applicant also belongs to one of two {\em groups}:
an {\em advantaged} group named $A$ or a
{\em disadvantaged} group named $D$.
(We will sometimes refer to the applicants from these 
groups as {\em $A$-applicants} and {\em $D$-applicants} respectively.)
The group membership of the applicant can be thought of as
a Boolean variable that we denote $\gp$, taking the value $A$ or $D$,
which gives the applicant an extended feature vector 
$(\x\^1, \ldots, \x\^k, \gp)$ with $k+1$ dimensions.
As a matter of notation, we will use 
$\x$, or sometimes a subscripted variable like $\x_i$, to denote a $k$-dimensional feature vector of the form
$(\x\^1, \ldots, \x\^k)$ (without the group membership variable $\gp$),
and we will use $\xb$ or $(\x, \gp)$ to denote an extended feature vector 
of the form $(\x\^1, \ldots, \x\^k, \gp)$.
Sometimes we will use $\x\^{k+1}$ to denote the group membership
variable $\gp$, so that the extended feature vector of an applicant can be
written $(\x\^1, \ldots, \x\^k, \x\^{k+1})$.

\xhdr{The productivity function}
Each applicant has a {\em productivity} that is a function of
their feature vector, and our goal is to admit applicants of high
productivity. 
In what follows, we don't impart a particular interpretation to
productivity except to say that we prefer applicants of higher
productivity; thus, productivity can correspond to whatever
criterion determines the true desired rank-ordering of applicants.
We write $f(\x, \gp)$ for the productivity
of an applicant with extended feature vector $(\x, \gp)$.
We will think of the values of $f$ as being specified by a 
look-up table, where each extended feature vector $(\x, \gp)$
is a {\em row} in the table;
we will therefore often refer to extended feature vectors as ``rows.''

We make the assumption that group membership has no effect on
productivity when the values of all the other features are fixed;
that is, for every $k$-dimensional feature vector $\x$, we have
$f(\x, A) = f(\x, D)$.
Thus, in a mild abuse of notation, we will sometimes write $f(\x)$
for this common value $f(\x, A) = f(\x, D)$.
We will also make the {\em genericity assumption} that
$f(\x) \neq f(\x')$ for distinct $k$-dimensional feature vectors $\x, \x'$.
(This is part of a broader genericity assumption that we state below.)

\xhdr{Measures and Disadvantage}
We now make precise the quantitative sense in which group $D$
experiences disadvantage relative to group $A$:
even though $f(\x, A) = f(\x, D)$ for all $\x$, 
a smaller fraction of group $D$
exhibits feature vectors $\x$ corresponding to larger (and hence
more desirable) values of the productivity $f$.
This is a natural way to think about disadvantage for our purposes:
conditional on the full set of features $\x$, group membership
has no effect on the value of $f$, but members of group $D$
in aggregate have fewer opportunities to obtain feature vectors
that produce large values of $f$.

We formalize this using the following definitions.
Let $\ms(\x,\gp)$ denote the fraction of the population whose
extended feature vector is equal to the row $(\x,\gp)$;
we will refer to this as the {\em measure} of row $(\x,\gp)$.
We will assume that every row has a positive measure, $\ms(\x,\gp) > 0$.
The disadvantage condition then states that
feature vectors yielding larger values of $f$ have higher representation
of group $A$:

\begin{stmt}
{\em (Disadvantage condition.)}
If $\x$ and $\x'$ are feature vectors 
such that $f(\x) > f(\x')$, then 
$$\frac{\ms(\x,A)}{\ms(\x,D)} > \frac{\ms(\x',A)}{\ms(\x',D)}.$$
\label{stmt:disad}
\end{stmt}

As one way to think about this formalization of disadvantage, we
can view the population fractions associated with each feature vector
as defining two distributions over possible values of $f$: 
a distribution over $f$-values for the advantaged group, 
and a distribution over $f$-values for the disadvantaged group.
The condition in \rf{stmt:disad} is equivalent to saying that
the distribution for the advantaged group exhibits what is known
as {\em likelihood-ratio dominance} with respect to the distribution
for the disadvantaged group
\cite{athey-monotone-likelihood,hopkins-ratio-orderings,milgrom-good-news,wolfstetter-microeconomics-book}; this is a standard way of formalizing
the notion that one distribution is weighted toward more advantageous
values relative to another, since it has increasingly high
representation at larger values.
It is interesting, however, to ask how much we might be able to
weaken the disadvantage condition and still obtain our main results;
we explore this question later in the paper, 
in Section \ref{subsec:disad}.

\xhdr{Averages}
There is another basic concept that will be useful in what follows:
taking the average value of $f$ over a set of rows.
This is defined simply as follows.
For a set of rows $S$, let $\ms(S)$ denote the total measure of all rows
in $S$: that is, $\ms(S) = \sum_{\xb \in S} \ms(\xb)$.
We then write $\avgf{f}{S}$ for the average value of
$f$ on applicants in rows of $S$; that is,
$$\avgf{f}{S} = \frac{ \sum_{\xb \in S} \ms(\xb) f(\xb) }{ \ms(S) }.$$

In terms of these average values, we can also state our full 
genericity assumption: 
beyond the condition $f(\x,A) = f(\x,D)$, there are no 
``coincidental'' equalities in the average values of $f$.

\begin{stmt}
{\em (Genericity assumption.)}
Let $S$ and $T$ be two distinct sets of rows such that if
$S = \{(\x,A)\}$ then $T \neq \{(\x,D)\}$.
Then $\avgf{f}{S} \neq \avgf{f}{T}$.
\label{stmt:genericity}
\end{stmt}

Note that this genericity assumption holds for straightforward reasons
if we think of all $f$-values as perturbed by random real numbers
drawn independently from an arbitrarily 
small interval $[-\eps,\eps]$.\footnote{To clarify one further point,
note that if $U$ is a set of $k$-dimensional feature vectors of
size greater than one, and $S = \{(\x,A) : \x \in U\}$ and 
$T = \{(\x,D) : \x \in U\}$, then 
$\avgf{f}{S} \neq \avgf{f}{T}$ follows purely from the 
disadvantage condition.}
The key point is that when the genericity condition does not hold, there is already some amount of ``simplification'' being performed by identities within $f$ itself; we want to study the process of simplification when --- as is typical in empirical applications --- $f$ is not providing such simplifying structure on its own.
(To take one extreme example of a failure of genericity, 
suppose the function $f$ didn't depend at all on one of the variables $\x\^{i}$; then we could clearly consider a version of the function that produced the same output without consulting the value of $\x\^{i}$, but this wouldn't in any real sense constitute a simplification of $f$.)

\subsection{Approximators}

We will consider admission rules that rank applicants and then
admit them in descending order.
One option would be to rank applicants by the value of $f$;
but as discussed above, there are many reasons why 
we may also want to work with a simpler approximation to $f$, 
ranking applicants by their values under this approximation.
We call such a function $g$ an {\em $f$-approximator}; it is defined
by specifying a partition of the applicant population into
a finite set of {\em cells} $C_1, C_2, \ldots, C_d$,
and approximating the value in each cell $C_i$
by a number $\gfn(C_i)$
equal to the average value of $f$ over the portion of the population
that lies in $C_i$.
Because we will be using the function $g$ to rank applicants
for admission, we will require that the cells of $g$ are sorted
in descending order:\footnote{We will allow distinct cells to have the
same value: $\gfn(C_i) = \gfn(C_j)$.
In an alternate formulation of the model, we could require that
all cells have distinct values; the main results would be essentially the same
in this case, although certain fine-grained details of the model's
behavior would change.}
$\gfn(C_i) \geq \gfn(C_j)$ for $i < j$.

A key point in our definition is that an $f$-approximator $g$
operates by simply specifying the partition into cells;
the {\em values} associated with these cells are determined
directly from the partition, as the average $f$-value in each cell.
This type of ``truth-telling'' constraint on the cell values
is consistent with our interest in studying the properties
of approximators as prediction functions.
Subsequent decisions that rely on an approximator $g$ could in 
principle post-process its values in multiple ways, but
our focus here --- as a logical underpinning for any such 
further set of questions ---
is on the values that such a function $g$ provides
as an approximation to the true function $f$.

\xhdr{Discrete $f$-approximators}
In the most basic type of $f$-approximator, 
each cell $C_i$ is a union of rows of the table defining $f$.
Thus, since each cell in an $f$-approximator receives
a value equal to the average value of $f$ over all
applicants in the cell, 
we assign cell $C_i$ the value
$\gfn(C_i) = \avgf{f}{C_i}$.

\xhdr{General $f$-approximators}
A fully general $f$-approximator can do more than this;
it can divide up individual
rows so that subsets of the row get placed in different cells.
We can imagine this taking place through randomization, or through
some other way of splitting the applicants in a single row.
For such a function $g$, we can still think of it as consisting
of cells $C_1, C_2, \ldots, C_d$, but these cells now have
continuous descriptions.  Thus, $g$ is described by a collection
of non-negative, non-zero vectors
$\phi_1, \phi_2, ..., \phi_d$, with each $\phi_i$
indexed by all the rows, and $\phi_i(\xb)$ specifying the total measure
of row $\xb$ that is assigned to the cell $C_i$ in $g$'s
partition of the space.
Thus we have the constraint $\sum_{i=1}^d \phi_i(\xb) = \ms(\xb)$,
specifying that each row has been partitioned.
We define $\ms(C_i)$ to be the total measure of all the fractions
of rows assigned to $C_i$; that is, $\ms(C_i) = \sum_{\xb} \phi_i(\xb)$.
The average value of the applicants in cell $C_i$ is given by 
$$\gfn(C_i) = \frac{ \sum_{\xb} \phi_i(\xb) f(\xb) }{ \ms(C_i) }.$$
An easy way to think about the approximator $g$ is that it assigns
a value to an applicant
with extended feature vector $\xb$ by mapping the applicant to cell $C_i$
with probability $\phi_i(\xb) / \ms(\xb)$, and then assigning them
the value $\gfn(C_i)$.

To prevent the space of $f$-approximators from containing 
functions with arbitrarily long descriptions, we will assume
that there is an absolute bound $\dimbd$ on the number of cells allowed
in an $f$-approximator.  (We will suppose that $\dimbd \geq 2^{k+1}$
so that a discrete $f$-approximator that puts each row of $f$
in a separate cell is allowed.)

It will also be useful to talk about the fraction of
applicants in cell $C_i$ that belong to each of the groups $A$ and $D$.
We write $\sb(C_i)$ for the fraction of applicants in $C_i$
belonging to group $D$; that is,
$$\sb(C_i) = \frac{ \sum_{(\x,D)} \phi_i(\x,D) }{ \ms(C_i) }.$$

Note that our more basic class of discrete $f$-approximators
$g$ that just partition
rows can be viewed as corresponding to a subset of this general
class of $f$-approximators as follows:
for all $\xb$, we simply require that
exactly one of the values $\phi_i(\xb)$ is non-zero.
In this case, note that $\gfn(C) = \avgf{f}{C}$ by definition.

Our notion of an approximator includes functions that use group membership;
that is, two applicants who differ only in group membership
(i.e. from respective rows $(\x,A)$ and $(\x,D)$ for some $\x$)
can be placed in different cells.
There are several reasons we allow this in our model.  
First, it is important to
remember that by construction we have assumed the true function
$f$ does not depend on group membership: $f(\x,A) = f(\x,D)$ for all $\x$. 
As a result, if we allowed a fully complex model that used
the true values $f(\x,\gp)$ for all applicants, there would be
no efficiency gains from using group membership.
Consequently, the main use of group membership in the constructions 
we consider is in fact to remediate the negative effects on group $D$
incurred through simplification.
Second, in many of our motivating applications, the distinction
between groups $A$ and $D$ does not correspond to a variable whose
use is legally prohibited; instead its use may be part of standard practice
for alleviating disadvantage.
For example, the distinction between $A$ and $D$ may correspond to
geographic disadvantage, or disadvantage based on some aspects of
past educational or employment history, all of which are dimensions
that are actively taken into account in admission-style decisions.
Finally, even in cases where group membership corresponds to
a legally prohibited variable, these prohibitions themselves are the result
of regulations that were put in place at least in part through
analysis that clarified the costs and benefits of allowing the use of
group membership.  As a result, to inform such decisions, it is standard
to adopt an analytical framework that allows for the use of such variables
ex ante, and to then use the framework to understand the consequences of 
prohibiting these variables.  In the present case, our analysis
will show how the use of these variables may be necessary to reduce
harms incurred through the application of simplified models.

At various points, it will also be useful to talk about approximators
that do not use group membership.  We formalize this as follows.

\begin{stmt}
We call a discrete $f$-approximator {\em group-agnostic} if 
for every feature vector $x = (\x\^{1}, \ldots, \x\^{k})$,
the two rows $(\x,A)$ and $(\x,D)$ belong to the same cell.
\label{stmt:def-group-ind}
\end{stmt}

\xhdr{Non-triviality and simplicity}
We say that a cell $C$ of an $f$-approximator is
{\em non-trivial} if it contains positive measure 
from rows $\xb, \xb'$ for which $f(\xb) \neq f(\xb')$; 
and we say that an $f$-approximator
itself is non-trivial if it has a non-trivial cell.

Any $f$-approximator that is discrete and non-trivial already
represents a form of simplification of $f$, in that it is
grouping together applicants with different $f$-values as part of
a single larger cell.
However, this is a very weak type of simplification, in that
the resulting $f$-approximator can still be relatively
lacking in structure.
We therefore focus in much of our analysis on a structured
form of simplicity that abstracts a key property of
most approaches to simplifying $f$.

Our core definition of simplicity is motivated by considering what are
arguably the most natural ways to construct 
collections of discrete, non-trivial $f$-approximators, as
special cases of one (or both) of the following definitions.
\begin{itemize}
\item {\em Variable selection.}
First, we could partition the rows into cells by projecting out
certain variables among $\x\^{1}, \ldots, \x\^{k}, x\^{k+1}$ 
(where again $x\^{k+1} = \gp$ is the group membership variable).
That is, for a set of indices $R \subseteq \{1, 2, \ldots, k+1\}$,
we declare two rows $\xb_i$ and $\xb_j$ to be equivalent if 
$\xb_i\^{\ell} = \xb_j\^{\ell}$
for all $\ell \in R$.
The cells $C_1, \ldots, C_d$ are then just the equivalence classes of
rows under this equivalence relation.
\item {\em Decision tree.}
Second, we could construct a partition of the rows using a decision
tree whose internal nodes consist of tests of the form
$\x\^{\ell} = b$ for $\ell \in \{1, 2, \ldots, k+1\}$, and $b \in \{0,1\}$.
(For $\x\^{k+1}$ we can use $0$ to denote $A$ and $1$ to denote $D$
in this structure.)
An applicant is mapped to a leaf of the tree using a standard
procedure for decision trees, in which they start at the root and
then proceed down the tree according to the outcome of the tests
at the internal nodes.
We declare two rows $\xb'$ and $\xb'$ to be equivalent if they are mapped
to the same leaf node of the decision tree, and again
define the cells $C_1, \ldots, C_d$ to be the equivalence classes of
rows under this relation.
\end{itemize}
We say that a cell $C_i$ is a {\em cube} if it consists of
all feature vectors obtained by specifying the values 
of certain variables and leaving the other variables unspecified.
That is, for some set of indices $R \subseteq \{1, 2, \ldots, k+1\}$,
and a fixed value $b_i \in \{0,1\}$ for each $i \in R$,
the cell $C_i$ consists of all vectors
$(\x\^1, \ldots, \x\^{k+1})$ for which $\x\^{i} = b_i$ for each $i \in R$.
We observe the following.

\begin{stmt}
For any discrete $f$-approximator constructed using
either variable selection or a decision tree,
each of its cells is a cube.
\label{stmt:cube}
\end{stmt}

We abstract these constructions into the notion of a
{\em simple $f$-approximator}.

\begin{stmt}
A {\em simple $f$-approximator} is a non-trivial
discrete $f$-approximator for which each cell is a cube.
\label{stmt:def-simple}
\end{stmt}

This is the basic definition of simplicity that we use in what follows.
If an $f$-approximator doesn't satisfy this definition, it must be
that at least one of its cells is obtained by gluing together
sets of rows that don't naturally align along dimensions in this sense;
this is the respect in which it is not {\em simple} in our framework.
At the end of this section, we will describe a generalization of
this definition that abstracts beyond the setting of Boolean functions,
and contains \rf{stmt:def-simple} as a special case.

We note that while decision trees provide a large, natural collection
of instances of simple $f$-approximators, there are still larger
collections of simple $f$-approximators that 
do not arise from the decision-tree construction described above.
To suggest how such further simple $f$-approximators can be obtained,
consider a partition of the eight rows associated 
with the three variables $\x\^1, \x\^2$, and the group membership
variable $\gp$: we define the cells to be 
$$
C_1 = \{(1,1,D)\}; 
C_2 = \{(1,1,A), (0,1,A)\}; 
C_3 = \{(0,1,D), (0,0,D)\}; 
$$
$$
C_4 = \{(1,0,A), (1,0,D)\};
C_5 = \{(0,0,A)\}.
$$
It is easy to verify that this $f$-approximator is simple, since each of
its five cells is a cube. But it cannot arise 
from the decision-tree construction described above
because no variable could be used for the test at the root of the tree:
$C_2$ contains rows with both values of $\x\^1$, while
$C_3$ contains rows with both values of $\x\^2$ and
$C_4$ contains rows with both values of $\gp$.

Our definition of simple approximators is not only motivated by
the natural generalization of methods including variable selection
and decision trees;
it also draws on basic definitions from behavioral science.
To the extent that the cells in an approximator represent the categories
or groupings of applicants that will be used by human decision-makers
in interpreting it, requiring that cells be cubes is consistent with
two fundamental ideas in psychology.
First, people mentally hold knowledge in categories where
like objects are grouped together;
such categories can be understood as
specifying some of the features but leaving others unspecified
\cite{mullainathan-categories,mullainathan-coarse-thinking,rosch1978cognition}. 
(For example, when we think of ``red cars,'' we have specified 
two values --- that the object is a car and the color is red --- 
but have left unspecified the age, size, manufacturer, and other features.)
This process of specifying some values and leaving others unspecified
is precisely our definition of a cube.
Second, the definition of a cube can be viewed equivalently as saying
that each cell is defined by a conjunction of conditions of
the form $x_i = b_i$.
Mental model theory from psychology emphasizes how conjunctive inferences
such as these are easier than disjunctive inferences
because they require only one mental model as opposed to a collection of
distinct models \cite{garcia2001conjunctive}.
(For example, ``red cars'' is a cognitively more natural concept for
human beings than the logically analogous concept,
``objects that are red or are cars.'')

\subsection{Admission Rules}

Any $f$-approximator $g$ creates an admission rule for applicants:
we sort all applicants by their value determined by $g$, and we admit them in
this order, up to a specified admission rate $r \in (0,1]$
that sets the fraction of the population we wish to admit.
Let $\adm_g(r)$ be the set of all applicants who are admitted under
the rule that sorts applicants according to $g$ and then admits
the top $r$ fraction under this order.

We can think about the sets $\adm_g(r)$ in terms of the ordering of the cells
of $g$ as $C_1, C_2 \ldots, C_d$ arranged
in decreasing order of $\gfn(C_i)$.
Let $r_j$ be the measure of the first $j$ cells in order, with $r_0 = 0$.
(We will sometimes write $r_j$ as $r_j\^{g}$ when we need to emphasize the dependence on $g$.)
Then for any admission rate $r$, the set of admitted applicants
$\adm_g(r)$ will consist of everyone in the cells $C_j$ for which 
$r_j \leq r$, along with a (possibly empty) portion of the next cell.
We can write this as follows:
if $j(r)$ is the unique index $j$ such that $r_{j-1} \leq r < r_j$,
then the set $\adm_g(r)$ consists of all the applicants in the
cells $C_1, C_2, \ldots, C_{j(r)-1}$, together with a
proper subset of $C_{j(r)}$.

\xhdr{Efficiency and equity}
Two key parameters of an admission rule are (i) its {\em efficiency},
equal to the average $f$-value of the admitted applicants; and
(ii) its {\em equity}, equal to the fraction of admitted applicants
who belong to group $D$.
Each of these is a function of $r$: 
the efficiency, denoted $\efc_g(r)$, is a decreasing
function of $r$ (since we admit applicants in decreasing order of
cell value), whereas the equity, denoted $\eqc_g(r)$,
can have a more complicated dependence
on $r$ (since successive cells may have a higher or lower representation
of applicants from group $D$).
We think of society's preferences as (at least weakly) favoring larger values for these two quantities 
(consistent with a social welfare approach to fairness and equity
\cite{KLMR-welfare-function}), but we will not impose any additional
assumptions on how efficiency and equity are incorporated into these
preferences.

We can write these efficiency and equity functions as follows.
First, let $\efd_g(r)$ be
the $f$-value of the marginal applicant admitted 
when the admission rate is $r$;
that is, $\efd_g(r) = \gfn(C_{j(r)})$.
Similarly, let $\eqd_g(r)$ be
the probability that the marginal applicant admitted
belongs to group $D$;
that is, $\eqd_g(r) = \sb(C_{j(r)})$.
Each of these functions is constant on the intervals of $r$ when
it is filling in the applicants from a fixed $C_j$;
that is, it is constant on each interval of the form 
$(r_{j-1},r_j)$, and it has a possible point of discontinuity
at points of the form $r_j$.

We can then write the efficiency and the equity simply as
the averages of these functions $\efd_g(\cdot)$ and $\eqd_g(\cdot)$
respectively:
$$\efc_g(r) = \frac{1}{r} \int_0^r \efd_g(t) ~ dt$$
and
$$\eqc_g(r) = \frac{1}{r} \int_0^r \eqd_g(t) ~ dt.$$
Note that even though $\efd_g(\cdot)$ and $\eqd_g(\cdot)$ have points of
discontinuity, the efficiency and equity functions are continuous.

\subsection{Improvability and Maximality}

Suppose we prefer admission rules with higher efficiency and higher equity,
and we are currently using an $f$-approximator $g$ with associated
admission rule $\adm_g(\cdot)$.
What would it mean to improve on this admission rule?
It would mean finding another $f$-approximator $h$ whose admission rule
$\adm_h(\cdot)$ produced efficiency and equity that were at least as good for
every admission rate $r$, and strictly better for at least one
admission rate $r^*$.
In this case, from the perspective of efficiency and equity, there
would be no reason not to use $h$ in place of $g$, 
since $h$ is always at least
as good, and sometimes better.

Formally, we will say that 
{\em $h$ weakly improves on $g$},
written $h \succeq g$,
if $\efc_h(r) \geq \efc_g(r)$ and $\eqc_h(r) \geq \eqc_g(r)$
for all $r \in (0,1]$.
We say that {\em $h$ strictly improves on $g$},
written $h \succ g$,
if $h$ weakly improves on $g$, and there exists an $r^* \in (0,1)$ for which
we have both $\efc_h(r^*) > \efc_g(r^*)$ and $\eqc_h(r^*) > \eqc_g(r^*)$.
(Viewing things from the other direction of the comparison,
we will write $g \preceq h$ if $h \succeq g$, and
$g \prec h$ if $h \succ g$.)

It can be directly verified from the definitions that the following
transitive properties hold.

\begin{stmt}
If $g_0$, $g_1$, and $g_2$ are $f$-approximators such that
$g_2 \succeq g_1$ and $g_1 \succeq g_0$, then $g_2 \succeq g_0$.
If additionally $g_2 \succ g_1$ or $g_1 \succ g_0$, then $g_2 \succ g_0$.
\label{stmt:improve-transitive}
\end{stmt}

We say that an $f$-approximator $g$ is {\em strictly improvable} if there
is an $f$-approximator $h$ that strictly improves on it.
For the reasons discussed above, 
it would be natural to favor $f$-approximators that are not
strictly improvable:
we say that an $f$-approximator $g$ is {\em maximal} if there
is no $f$-approximator $h$ that strictly improves on it.

The set of maximal $f$-approximators is in general quite rich
in structure, but we can easily pin down perhaps 
the most natural class of examples: 
if we recall that a {\em trivial} approximator is one that never
combines applicants of distinct $f$-values into the same cell, then
it is straightforward to verify that 
every trivial $f$-approximator $g$ is maximal, simply because there
cannot be any $f$-approximator $h$ and admission rate $r^*$ for which
$\efc_h(r^*) > \efc_g(r^*)$.

\begin{stmt}
Every trivial $f$-approximator is maximal.
\label{stmt:trivial-maximal}
\end{stmt}

\rf{stmt:trivial-maximal} establishes the existence of maximal
$f$-approximators, but we can say more about them via
the following fact, which establishes that every $f$-approximator
has at least one maximal approximator ``above'' it.

\begin{stmt}
For every $f$-approximator $g$, there exists a maximal $f$-approximator
$h$ that weakly improves it.
\label{stmt:maximal-improve}
\end{stmt}

Since the proof of \rf{stmt:maximal-improve} is fairly technical,
and the methods used are not needed in what follows,
we defer the proof to the appendix.

\xhdr{Improvability in Efficiency and Equity}
At various points, it will be useful to talk about 
pairs of approximators that satisfy the definition of
strict improvability only for efficiency, or only for equity.

Thus, for two $f$-approximators $g$ and $h$, will say that
$h$ {\em strictly improves $g$ in efficiency}
if at every admission rate $r$, the average $f$-value of
the applicants admitted using $h$ is at least as high as the
average $f$-value of the applicants admitted using $g$,
and it is strictly higher for at least one value of $r$.
We will write this as $h \succ_\efd g$ or equivalently $g \prec_\efd h$;
in the notation developed above, it means that
$\efc_h(r) \geq \efc_g(r)$ for all $r \in (0,1]$, and
$\efc_h(r) > \efc_g(r)$ for at least one value of $r$.
Correspondingly, we will say that $h$ {\em strictly improves $g$ in equity},
written $h \succ_\eqd g$ or equivalently $g \prec_\eqd h$, if
$\eqc_h(r) \geq \eqc_g(r)$ for all $r \in (0,1]$, and
$\eqc_h(r) > \eqc_g(r)$ for at least one value of $r$.
We observe that the analogue of our fact about transitivity,
\rf{stmt:improve-transitive}, holds for both $\succ_\efd$ and
$\succ_\eqd$.


\subsection{Main Results}
\label{subsec:model-results}

Given the model and definitions developed thus far, it is easy
to state the basic forms of our two main results.
We let $f$ be an arbitrary function over a set of 
extended feature vectors for which the disadvantage condition
\rf{stmt:disad} and genericity assumption \rf{stmt:genericity} hold.

\xhdr{First Result: Simple Functions are Improvable}
In Section~\ref{sec:main-result}, we will prove the following result.

\begin{stmt}
Every simple $f$-approximator is strictly improvable.
\label{stmt:simple-improvable}
\end{stmt}

This result expresses the crux of the tension between simplicity
and equity --- for every admission rule based on a simple
$f$-approximator, we can find another admission rule that is at least
as good for every admission rate, and which for some admission rates
strictly improves on it in both efficiency {\em and} equity.
Thus, whatever one's preferences are for efficiency and equity,
this alternate admission rule should be favored on these two grounds.

We will prove this result in Section \ref{sec:main-result}.
To get a sense for one of the central ideas in the proof, it is useful to
consider a simple illustrative special case of the result:
if $g$ is the $f$-approximator that puts all applicants into a 
single cell $C$, how do we strictly improve it?

We can construct a strict improvement on $g$ as follows.
First, for the approximator $g$, note that the function $\efc_g(r)$ is a
constant, independent of $r$ and equal to the average $f$-value
over the full population of applicants.
The function $\eqc_g(r)$ is also a constant, equal to the fraction
of $D$-applicants in the full population.
We construct a strict improvement on $g$ by first finding a row
associated with group $D$ that has an above-average $f$-value 
(such a row exists since $f$ is not a constant function and
doesn't depend on group membership)
and pulling this row into a separate cell that we can admit first.
Specifically, let $\x$ be the feature vector for which $f(\x,D)$ is maximum,
and consider the approximator $h$ consisting of two cells:
$C_1$ containing just the row $(x,D)$, and $C_2$ containing 
all other rows.
The function $\efc_h(r)$ is equal to $f(\x,D)$ for $r \leq \ms(\x,D)$,
and then it decreases linearly to $\efc_g(1)$.
The function $\eqc_h(r)$ is equal to $1$ for $r \leq \ms(\x,D)$,
and then it decreases linearly to $\eqc_g(1)$.
It follows that $h$ strictly improves on $g$.
 
In the full proof of the result, we will need several different
strategies for pulling rows out of a cell so as to produce
an improvement. In addition to pulling out rows of high $f$-value
associated with group $D$, we will also sometimes need to pull out rows
of low $f$-value associated with group $A$;
and sometimes we will need to pull out just a fraction of a row,
producing a non-discrete approximator.
The crux of the proof will be to show that some such operation
is always possible for a simple approximator.

\xhdr{Second Result: Simplicity Can Transform Disadvantage into Bias}
The second of our main results concerns group-agnostic
approximators.
Suppose that $g$ is a group-agnostic $f$-approximator, so
that rows of the form $(\x,A)$ and $(\x,D)$ always appear
together in the same cell.
Perhaps the most basic example of such a structure is the
unique $f$-approximator $g^\circ$ that is both group-agnostic and trivial:
it consists of $2^k$ cells, each consisting of the two rows
$\{(\x,A), (\x,D)\}$ for distinct feature vectors $\x$.
Since $f(\x,A) = f(\x,D)$ for all feature vectors $\x$, this
approximator $g^\circ$ has the property that it would not be 
strictly improved in efficiency if we were to split each cell
$\{(\x,A), (\x,D)\}$ into two distinct cells, one with each row,
since these two new smaller cells would each have the same value.

Now, however, consider any group-agnostic $f$-approximator $g$ that is
non-trivial, in that it has cells containing rows of different $f$-values.
(Recall that group-agnostic approximators are by definition discrete,
in that each row is assigned in its entirety to a cell rather than
being split over multiple cells.)
Let $\gpdep(g)$ be the $f$-approximator that we obtain from $g$ by
splitting each of its cells $C_i$ into two sets according to group 
membership --- that is, into the two cells
$\{(\x,A) : (\x,A) \in C_i\}$ and 
$\{(\x,D) : (\x,D) \in C_i\}$ ---
and then merging cells of the same $\gfn$-value.

In Section \ref{sec:transform}, we will show that
as long as $g$ is non-trivial, this operation strictly improves efficiency,
and strictly worsens equity:

\begin{stmt}
If $g$ is any non-trivial group-agnostic $f$-approximator, 
then $\gpdep(g)$ strictly improves $g$ in efficiency, and
$g$ strictly improves $\gpdep(g)$ in equity.
\label{stmt:group-split}
\end{stmt}

This result highlights a key potential concern that arises when
we approximate a productivity function $f$ in the presence of disadvantage.
Consider a decision-maker who is interested in maximizing
efficiency, and does not have preferences about equity.
When they are using the true $f$-values for each applicant,
as $g^\circ$ does above, there is no incentive for this
decision-maker to take group membership into account.
But as soon as they are using any 
non-trivial group-agnostic approximator $g$, 
there becomes an incentive to incorporate knowledge of group membership,
since splitting the cells of $g$ according to group membership 
in order to produce $\gpdep(g)$ will create a strict improvement
in efficiency.
However, this operation comes at a cost to the disadvantaged group $D$,
since $g$ strictly improves $\gpdep(g)$ in equity.

Thus, any non-trivial group-agnostic approximation to $f$ is effectively
transforming disadvantage into bias: where the decision-maker
was initially indifferent to group membership, the process of
suppressing information so as to approximate $f$ created an 
incentive to use a rule that is explicitly biased in using
group membership as part of the decision.

\xhdr{Comparing Different Forms of Simplicity}
It is useful to observe that our two results 
\rf{stmt:simple-improvable} and \rf{stmt:group-split} are
both based on simplifying the underlying function $f$,
but in different ways.
The first is concerned with approximators that are {\em simple}
in the sense of \rf{stmt:def-simple},
that each cell is obtained by fixing the values of
certain variables $\x\^{i}$ and leaving the others unrestricted.
The second is concerned with approximators that are 
{\em group-agnostic}, in the sense that rows of the form
$(\x,A)$ and $(\x,D)$ always go into the same cell;
but it applies to any non-trivial group-agnostic approximator.

Before proceeding to some illustrative examples and to the proofs
of these results,
we first cast them in a more general form.

\subsection{A More General Formulation}

It turns out that the proof technique we use for our main results
can be used to establish a corresponding pair of
statements in a more general model.
It is worth spelling out this more general version, since it makes
clear that our results do not depend on a model in which the 
feature vectors must be comprised of $k$ Boolean coordinates;
in fact, all that matters is that there is an arbitrary
finite set of feature vectors.

We define this more general formulation as follows.
Suppose that each individual is described by one of $n$ possible
feature vectors, labeled $\x_1, \x_2, \ldots, \x_n$, along with 
a group membership variable $\gp$ which, as before, 
can take the value $A$ or $D$.
As before, the fraction of the population described 
by the extended feature vector $(\x_i,\gp)$
is given by $\ms(\x_i,\gp)$; the 
productivity of an individual described by $(\x_i,\gp)$ is
given by a function $f(\x_i,\gp)$; and group membership
has no effect on $f$ once we know the value of $\x_i$: that is,
$f(\x_i,A) = f(\x_i,D)$, and we will refer to both as $f(\x_i)$.
We will continue to refer to each extended feature vector
$(\x,\gp)$ as a {\em row} $\xb$ (of the look-up table defining $f$),
and assume that $f(\x_i,\gp) \neq f(\x_j,\gp')$ 
for different feature vectors $\x_i, \x_j$; for convenience
we will index the feature vectors $\x_1, \x_2, \ldots, \x_n$,
so that $f(\x_j) > f(\x_i)$ when $j > i$.
The disadvantage condition also remains essentially the same as before:
if $\x_i$ and $\x_j$ are feature vectors such that 
$f(\x_j) > f(\x_i)$, then 
$$\frac{\ms(\x_j,A)}{\ms(\x_j,D)} > \frac{\ms(\x_i,A)}{\ms(\x_i,D)}.$$

To see that our original Boolean model is a special case of this more general one,
simply set $n = 2^k$ and let $\x_1, \x_2, \ldots, \x_n$ be the
$n$ possible vectors consisting of $k$ Boolean values, sorted 
in increasing order of $f$-value.
(That is, each feature vector $x_i$ has the form $(x_i\^{1}, \ldots, x_i\^{k})$ for Boolean variables $x_i\^{1}, \ldots, x_i\^{k}$).
The remainder of the model is formulated in exactly the same way as before,
with one exception: the definition of a simple $f$-approximator was
expressed in terms of the Boolean coordinates of the feature vectors
(as part of the definition of a {\em cube}), and so we need to generalize
this definition to our new setting, resulting in a class of
approximators that contains more than just simple ones.

\xhdr{Graded approximators}
To motivate our generalization of simple approximators, which we will refer
to as {\em graded approximators}, we begin with some notation.
For a cell $C_i$ in a discrete $f$-approximator, let
$C_i\^{A}$ denote the set of feature vectors $\x$ such that
$(\x,A)$ is a row of $C_i$, and let
$C_i\^{D}$ denote the set of feature vectors $\x$ such that
$(\x,D)$ is a row of $C_i$.
We observe that a simple $f$-approximator has the property
that for every cell $C_i$, either one of $C_i\^{A}$ or $C_i\^{D}$
is empty, or else $C_i\^{A} = C_i\^{D}$.
Thus we have
$C_i\^{A} \subseteq C_i\^{D}$ or $C_i\^{D} \subseteq C_i\^{A}$ for all cells.

We take this condition as the basis for our definition of
graded approximators.

\begin{stmt}
A {\em graded $f$-approximator} is a non-trivial
discrete $f$-approximator
whose cells $C_1, C_2, \ldots, C_d$ satisfy
$C_i\^{A} \subseteq C_i\^{D}$
or $C_i\^{D} \subseteq C_i\^{A}$ for each $i$.
In the special case when 
the feature vectors are comprised of Boolean coordinates,
all simple $f$-approximators are graded.
\label{stmt:def-graded}
\end{stmt}

The more general formulation of our first 
result applies to graded approximators. 
Since all simple appoximators are graded,
this more general version thus extends the earlier formulation
\rf{stmt:simple-improvable}.
We state the result as follows,
given an arbitrary function $f$
for which the disadvantage condition
and genericity assumption 
(the analogues of \rf{stmt:disad} and \rf{stmt:genericity}) hold.

\begin{stmt}
Every graded $f$-approximator is strictly improvable.
\label{stmt:graded-improvable}
\end{stmt}

For the second result, we note that the definition of a
group-agnostic approximator remains the same in this more
general model --- that for every feature vector $x_i$, the two
rows $(x_i,A)$ and $(x_i,D)$ should belong to the same cell ---
and so our second result continues to have the same statement
as in \rf{stmt:group-split}.
It is also worth observing that every group-agnostic $f$-approximator
in our more general model is {\em graded} in the sense of 
\rf{stmt:def-graded}, since $C_i\^{A} = C_i\^{D}$ for every cell
in a group-agnostic approximator by definition.

\begin{figure}[t]
\begin{center}
\begin{tabular}{@{}|p{0.25in}|p{0.25in}|p{0.25in}|p{0.25in}|p{0.5in}|}
\hline
$x\^{1}$ & $x\^{2}$ & $\gp$ & $f$ & $\ms$
\\\hline
1 & 1 & $D$ & 1 & $q_1 q_2' /2$
\\\hline
1 & 1 & $A$ & 1 & $p_1 p_2 /2$
\\\hline
1 & 0 & $D$ & $\fval_{10}$ & $q_1 p_2' /2$
\\\hline
1 & 0 & $A$ & $\fval_{10}$ & $p_1 q_2 /2$
\\\hline
0 & 1 & $D$ & $\fval_{01}$ & $p_1 q_2' /2$
\\\hline
0 & 1 & $A$ & $\fval_{01}$ & $q_1 p_2 /2$
\\\hline
0 & 0 & $D$ & 0 & $p_1 p_2' /2$
\\\hline
0 & 0 & $A$ & 0 & $q_1 q_2 /2$
\\\hline
\end{tabular}
\caption{An example of a function with two Boolean variables and a
group membership variable.
\label{fig:example}
}
\end{center}
\end{figure}

\section{Examples and Basic Phenomena}
\label{sec:examples}

To make the model and definitions more concrete, it is useful
to work out an extended example; in the process we will also
identify some of the model's basic phenomena.
For purposes of this example, 
we will make use of the initial Boolean formulation
of our model, rather than the generalization to graded approximators.

In our example, there are two Boolean variables $x\^{1}$ and $x\^{2}$, along
with the group membership $\gp$.
Applicants have much higher productivity when $x\^{1} = x\^{2} = 1$ than
for any other setting of the variables; we define
$f(1,1) = 1$ and $f(0,0) = 0$; and we define $f(1,0) = \fval_{10}$ and 
$f(0,1) = \fval_{01}$ for very small 
numbers $\fval_{10} > \fval_{01} > 0$.
By choosing $\fval_{10}$ and $\fval_{01}$ appropriately (for example,
uniformly at random from a small interval just above $0$)
it is easy to ensure that the genericity condition holds.

Half the population belong to the advantaged group $A$ and
the other half belongs to the disadvantaged group $D$.
To define the distribution of values in each group,
we fix numbers $p_1 > p_2$ between $0$ and $1$, close enough to $1$ that
$p_1 p_2 > \frac12$.
For compactness in notation, we will write $q_i$ for $1 - p_i$.
In the advantaged group $A$, each applicant has $x_i = 1$ independently
(for $i = 1, 2$) with probability $p_i$, and $x_i = 0$ otherwise.
In the disadvantaged group $D$, the situation is (approximately) reversed.
Each applicant in group $D$ has $x_1 = 1$ independently
with probability $q_1 = 1-p_1$, and $x_1 = 0$ otherwise.
To ensure the genericity condition, we choose a value $p_2'$
very slightly above $p_2$ (but smaller than $p_1$), 
and define $q_2' = 1 - p_2'$;
each applicant in group $D$ has $x_2 = 1$ independently
with probability $q_2'$, and $x_2 = 0$ otherwise.
The full function $f$ is shown in Figure \ref{fig:example};
it is easy to verify that the disadvantage condition \rf{stmt:disad} holds
for this example.

\subsection{Two Trivial Approximators}

Before discussing simple approximators, it is worth briefly remarking
on two natural trivial $f$-approximators.
The first, which we will denote $g^*$, puts each row into a
single cell, and it sorts these eight cells in the order given by
reading the table in Figure \ref{fig:example} from top to bottom.
The second, $g^\circ$, which was discussed in 
Section \ref{subsec:model-results}, groups together pairs
of rows that differ only in the group membership variable $\gp$:
thus it puts the first and second row into a cell $C_1$;
then the third and fourth into a cell $C_2$; and so on, producing four cells.

Even the comparison between these two trivial approximators highlights
an important aspect of our definitions.
$g^*$ and $g^\circ$ have the same efficiency functions 
$\efc_{g^*}(\cdot)$ and $\efc_{g^\circ}(\cdot)$.
The equity function of $g^*$, on the other hand, is clearly preferable
to the equity function of $g^\circ$: we have 
$\eqc_{g^*}(r) \geq \eqc_{g^\circ}(r)$ for all $r$, and
$\eqc_{g^*}(r) > \eqc_{g^\circ}(r)$ for a subset of values of $r$.
It is not the case that $g^*$ strictly improves on $g^\circ$ according
to our definitions, however, since that would require the existence
of an $r$ for which 
$\efc_{g^*}(r) > \efc_{g^\circ}(r)$ {\em and} $\eqc_{g^*}(r) > \eqc_{g^\circ}(r)$.
This is clearly not possible, since $\efc_{g^*}(\cdot)$ and $\efc_{g^\circ}(\cdot)$
are the same function.
In fact, both $g^*$ and $g^\circ$ are maximal.
The issue is that our definition of strict improvement sets a
very strong standard: one admission rule strictly improves another
if and only if it is better for a decision-maker who cares only about
efficiency, or only about equity, or about any combination of the two.
And from the point of view of a decision-maker who cares only
about efficiency, $g^*$ is not strictly better than $g^\circ$.

This is a crucial point; we have chosen a deliberately strong
standard for the definition of strict improvement (and a weak 
definition of maximality) because it allows us to state our result
\rf{stmt:simple-improvable} in a correspondingly strong way:
even though it requires a lot to assert that an approximator $h$
strictly improves an approximator $g$, it is nevertheless the case
that every simple approximator is strictly improvable.

\begin{figure}[t]
\begin{center}
\includegraphics[width=.90\textwidth]{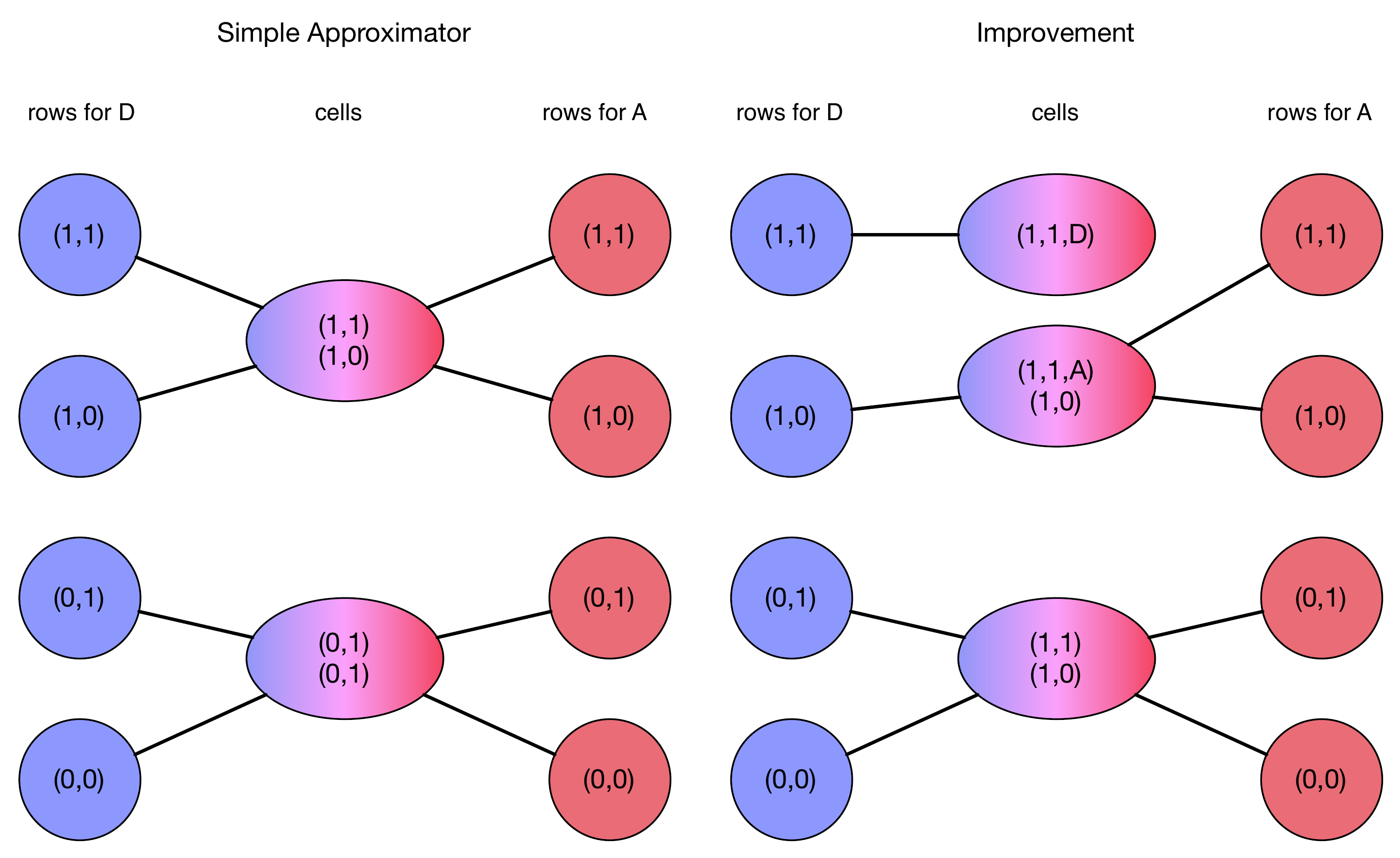}
\caption{
On the left, we have a depiction of the approximator $g_1$, which
simply groups applicants by the value of $\x\^{1}$:
the blue and red circles represent the rows associated with
$D$ and $A$ respectively, and they are linked to ovals 
representing the cells that contain them.
We can improve both the efficiency and the equity of $g_1$ if we
use the construction on the right, creating
a new cell that pulls out the row $(1,1,D)$ --- 
containing those applicants
from group $D$ with $\x\^{1} = \x\^{2} = 1$ --- and ranking it separately.
\label{fig:thm1}
}
\end{center}
\end{figure}

\subsection{Improving a Simple Approximator}
\label{sec:example-improve}

Now, let's consider a natural example of a simple $f$-approximator, and
see how it can be strictly improved.
The approximator we consider is $g_1$, which creates two cells:
$C_1$, consisting of all applicants for whom $x\^{1} = 1$, and 
$C_2$, consisting of all applicants for whom $x\^{1} = 0$.
Intuitively, this corresponds to projecting the applicants onto
just the variable $x\^{1}$, ignoring the values of $x\^{2}$ and 
the group membership $\gp$.

For $g_1$, 
we have $\ms(C_1) = p_1/2 + q_1/2 = 1/2$, and $\ms(C_2) = 1/2$ as well.
In this first cell $C_1$, the fraction of $D$-applicants is 
$$\frac{q_1/2}{p_1/2 + q_1/2} = q_1,$$
and so $\eqc_{g_1}(r) = q_1$ for all $r \leq 1/2$.
The average $f$-value of the applicants in $C_1$, working directly
from the definition, is
$p_1 p_2 + q_1 q_2' + \fval_{10}(p_1 q_2 + q_1 p_2')$,
which is close to $p_1 p_2 + q_1 q_2'$ since $\fval_{10}$ is very small.
This is the value of $\efc_{g_1}(r)$ for all $r \leq 1/2$.

Now, let's look for a function that strictly improves $g_1$.
A natural first candidate to consider is $g^*$: its efficiency
cannot be improved at any admission rate $r$
since it orders all applicants in decreasing order
of $f$-value; and subject to this (i.e. as a tie-breaker)
it puts $D$-applicants ahead of $A$-applicants.
It turns out, though, that there are values of the admission rate $r$
for which $g_1$ has better equity than $g^*$.
In particular, consider $r^* = (p_1 p_2 + q_1 q_2')/2$, when 
the set $\adm_{g^*}(r^*)$ admitted according to $g^*$ is precisely
those applicants with $x\^{1} = x\^{2} = 1$.
For this value of $r^*$, we have
$$\eqc_{g^*}(r^*) = \frac{q_1 q_2'}{p_1 p_2 + q_1 q_2'}.$$
But since 
$r^* = (p_1 p_2 + q_1 q_2')/2 < (p_1 + q_1)(p_2 + q_2')/2 < (p_1 + q_1)(p_2 + q_2)/2 = 1/2$, 
we have $$\eqc_{g_1}(r^*) = q_1 = \frac{q_1 q_2'}{p_1 q_2' + q_1 q_2'} > 
\frac{q_1 q_2'}{p_1 p_2 + q_1 q_2'} = \eqc_{g^*}(r^*),$$
and therefore $g^*$ does not strictly improve on $g_1$.

Arguably, this calculation implicitly connects to the 
qualitative intuition that simpler rules may be fairer in general:
since the distributions of both $x\^{1}$ and $x\^{2}$ confer disadvantage
on group $D$, by using only one of them rather than both
(thus using $g_1$ instead $g^*$) we give up some efficiency but we
are rewarded by improving equity at a crucial value of the admission rate:
specifically, the rate $r^*$ corresponding to the fraction of 
``top applicants'' (of $f$-value equal to 1) in the population.

But this intuition is misleading, because in fact there
are approximators that improve on $g_1$ in both efficiency
{\em and} equity; it's just that $g^*$ isn't one of them.
An approximator that we can use is $h$, which starts from
$g_1$ and then splits the cell $C_1$ into two cells:
$C_1'$, consisting of applicants from row $(1,1,D)$, and
$C_1''$, consisting of applicants from the three rows
$(1,1,A), (1,0,A)$, and $(1,0,D)$.
The approximator $h$ thus has the three cells $C_1', C_1'', C_2$,
in this order.
We can now check that $h$ is at least as good as $g_1$ at every
admission rate $r$, since it is admitting the applicants of cell $C_1$
in a subdivided order --- everyone in $C_1'$ followed by 
everyone in $C_1''$ --- and the applicants in $C_1'$ have a higher
average $f$-value than the applicants of $C_1$, and all of them
belong to group $D$.
Moreover, this means that when $r = \ms(C_1')$, 
we have $\efc_{h}(r) > \efc_{g_1}(r)$ and $\eqc_{h}(r) > \eqc_{g_1}(r)$,
and hence $h$ strictly improves $g_1$.

Figure \ref{fig:thm1} shows schematically how we produce 
$h$ from $g_1$.
Initially, $g_1$ groups all the rows with $\x\^{1} = 1$ into 
one cell, and all the rows with $\x\^{1} = 0$ into another.
We then produce $h$ by pulling the row $(1,1,D)$ out of this 
first cell and turning it into a cell on its own, with both a higher
average $f$-value and a positive contribution to the equity.

This example gives a specific instance of the general construction
that we will use in proving our first
main result \rf{stmt:graded-improvable}: breaking apart a
non-trivial cell so as to admit a subset with both a higher average
$f$-value and a higher representation of $D$-applicants.  This construction
also connects directly to both our earlier discussion of improving
simple approximators, and to the line of intuition expressed in the
introduction --- that simplifying by suppressing variables can prevent
the strongest disadvantaged applicants from demonstrating their
strength.

\begin{figure}[t]
\begin{center}
\subfigure[\emph{Approximator $g_1$ using only $x\^{1}$}]{
\begin{tabular}{@{}|p{0.25in}|p{0.25in}|p{0.25in}|p{2.25in}|p{0.5in}|}
\hline
$x\^{1}$ & $x\^{2}$ & $\gp$ & $g_1$ & $\ms$
\\\hline
1 & any & any & $p_1 p_2 + q_1 q_2' + \fval_{10}(p_1 q_2 + q_1 p_2')$ & 1/2
\\\hline
0 & any & any & $\fval_{01}(p_1 q_2' + q_1 p_2)$ & 1/2
\\\hline
\end{tabular}
\label{subfig:group-ind}
}
{~} \\
\subfigure[\emph{Approximator $\gpdep(g_1)$ using $x\^{1}$ and $\gp$}]{
\begin{tabular}{@{}|p{0.25in}|p{0.25in}|p{0.25in}|p{2.25in}|p{0.5in}|}
\hline
$x\^{1}$ & $x\^{2}$ & $\gp$ & $\gpdep(g_1)$ & $\ms$
\\\hline
1 & any & $A$ & $p_2  + \fval_{10} q_2$  & $p_1/2$
\\\hline
1 & any & $D$ & $q_2' + \fval_{10} p_2'$ & $q_1/2$
\\\hline
0 & any & $A$ & $\fval_{01} p_2$  & $q_1/2$
\\\hline
0 & any & $D$ & $\fval_{01} q_2'$ & $p_1/2$
\\\hline
\end{tabular}
\label{subfig:group-dep}
}
\caption{The $f$-approximator $g_1$ has only two cells, based on
the value of $\x\^{1}$.  When we split each of these cells by
using the value of the group membership variable $\gp$ as well,
we end up with an $f$-approximator $\gpdep(g_1)$ that is strictly
better in efficiency and strictly worse in equity.
Thus, for a decision-maker interested in maximizing efficiency,
the suppression of $\x\^{2}$ leads to an incentive to consult
the value of group membership, in a way that reduces equity for group $D$.
\label{fig:group-split}
}
\end{center}
\end{figure}

\begin{figure}[t]
\begin{center}
\includegraphics[width=.94\textwidth]{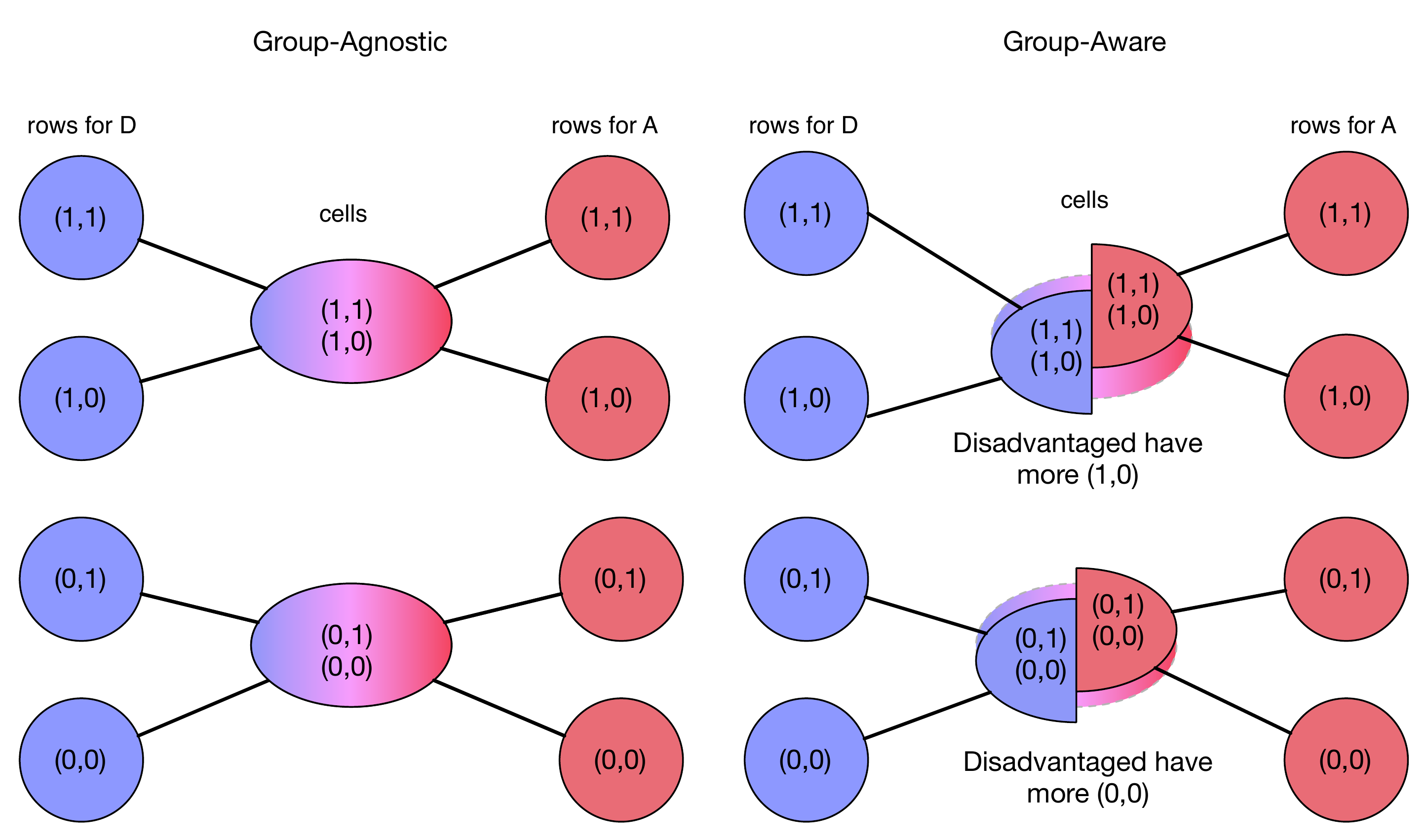}
\caption{
As in Figure \ref{fig:thm1}, the approximator on the left is $g_1$,
which groups applicants by the value of $\x\^{1}$.
If we split each of the two cells using group membership,
we get the approximator $\gpdep(g_1)$ on the right, with four cells in total: two associated with group $A$ and two associated with group $D$.  
In $\gpdep(g_1)$,
the cells associated with $A$ have moved slightly upward in value,
and the cells associated with $D$ have moved slightly downward in value.
As a result, this new approximator $\gpdep(g_1)$ is more
efficient but less equitable than the original group-agnostic
approximator $g_1$.
\label{fig:thm2}
}
\end{center}
\end{figure}

\subsection{Adding Group Membership to an Approximator}

Because the approximator $g_1$ is group-agnostic, we can also
use it to provide an example of the effect we see when we move
from a group-agnostic approximator $g$ to the 
version $\gpdep(g)$ in which we split each of $g$'s cells using
group membership.

Specifically, consider the cells of $g_1$, denoted 
$C_1$ and $C_2$ in the previous subsection.
$C_1$ groups together the four rows in which $\x\^{1} = 1$, and
$C_2$ groups together the four rows in which $\x\^{1} = 0$.
Now, suppose that we split $C_1$ into two cells:
$C_1'$ consisting of the two rows in which $\x\^{1} = 1$ and $\gp = A$, and
$C_1''$ consisting of the two rows in which $\x\^{1} = 1$ and $\gp = D$.
Working from the definitions, the average $f$-value of an applicant
in $C_1'$ is $p_2 + \fval_{10} q_2$, 
while the average $f$-value of an applicant
in $C_1''$ is $q_2' + \fval_{10} p_2'$.
Similarly, if we split $C_2$ into cells 
$C_2'$ with $\x\^{1} = 1$ and $\gp = A$, and 
$C_2''$ with $\x\^{1} = 1$ and $\gp = D$, then
the average $f$-value of an applicant
in $C_2'$ is $\fval_{01} p_2$, 
while the average $f$-value of an applicant
in $C_2''$ is $\fval_{01} q_2'$.

The pair of tables in Figure \ref{fig:group-split} provides one
way of summarizing these calculations: each row represents a
cell in which certain variables are fixed and others are set to
``any,'' meaning that the cell averages over rows with any value
allowed for these variables.
To go from $g_1$ to $\gpdep(g_1)$, we convert the first row in 
the first table into the top two rows in the second table, and
we convert the second row in 
the first table into the bottom two rows in the second table.

We have the sequence of inequalities
$$\gfn(C_1') > \gfn(C_1) > \gfn(C_1'') 
   > \gfn(C_2') > \gfn(C_2) > \gfn(C_2''),$$
reflecting the fact that using group membership in conjunction with 
$\x\^{1}$ results in a partition of each cell into a subset
with higher average $f$-value and a subset with lower average $f$-value.
Since $\gpdep(g_1)$ consists of the cells
$C_1', C_1'', C_2', C_2''$, it strictly improves $g_1$ in efficiency.
But since $\gpdep(g)$ places rows associated with group $A$ ahead
of the corresponding rows associated with group $D$, it follows that
$g$ strictly improves $\gpdep(g)$ in equity.

Figure \ref{fig:thm2} provides another way to depict
the transformation from $g_1$ to $\chi(g_1)$: when we split
the cells of $g_1$ using group membership, the cells associated with
group $A$ move slightly upward and the cells associated with
group $D$ move slightly downward, producing both the increase in
efficiency and the reduction in equity.

Thus, for a decision-maker who wants to maximize efficiency,
the $f$-approximator $g_1$ creates an incentive to consult the
value of $\gp$ encoding group membership, since doing so 
leads to a strict improvement in efficiency.
The resulting rule $\gpdep(g_1)$, 
however, is explicitly biased against applicants from group $D$,
in that it uses group membership information and results in
reduced equity for group $D$.
This effect wouldn't have happened had we started from the
$f$-approximator $g^{\circ}$ that uses the values of both
$\x\^{1}$ and $\x\^{2}$; 
in that case, efficiency would not be improved 
by using group membership information.
It is by suppressing information about the value of
$\x\^{2}$ that $g_1$ creates an incentive to incorporate
group membership.

\section{Proof of First Result: Simple Functions are Improvable}
\label{sec:main-result}

In this section, we prove our first main result in its general form,
\rf{stmt:graded-improvable}.
The basic strategy will be an extension of the idea used in
the discussion after the statement of \rf{stmt:simple-improvable} and in 
the example from Section \ref{sec:example-improve}: given a
simple (or graded) $f$-approximator $g$, we will show how to 
break up one or more of its cells, changing the order in which 
rows are admitted, so that the efficiency and equity don't decrease,
and for some admission rate we are admitting applicants of higher
average $f$-value and with a greater fraction of $D$-applicants.

It will turn out that the most challenging case is 
when we have an $f$-approximator $g$ in which each non-trivial 
cell consists entirely of rows associated with $A$
or entirely of rows associated with $D$.
We will call such an approximator {\em separable} (since
its non-trivial cells separate the two groups completely);
it is easy to verify from the definition of graded approximators
in \rf{stmt:def-graded} that every separable approximator is graded.
For this case, we will need to first prove a preliminary combinatorial lemma, 
which in turn draws on a consequence of the disadvantage condition.

We will work in the general model, where we have 
an arbitrary set of feature vectors $\{\x_1, \x_2, \ldots, \x_n\}$,
indexed so that $f(\x_i) < f(\x_j)$ for $i < j$;
this results in a set of $2n$ rows of the form $(\x_i, \gp)$
for a group membership variable $\gp$.
We will use $y_i$ to denote $f(\x_i)$, so the set of possible $f$-values
is $\{y_1, \ldots, y_n\}$.

\subsection{A Consequence of the Disadvantage Condition}

It is not hard to show that the disadvantage condition implies
that the average $f$-value over the $A$-applicants is higher than
the average $f$-value over the $D$-applicants.
But we would like to establish something stronger, as follows.
For a set of feature vectors $S \subseteq \{\x_1, \x_2, \ldots, \x_n\}$,
let $S_A = \{(x_i, A): \x_i \in S\}$ and
$S_D = \{(x_i, D): \x_i \in S\}$.
For a set $S$, we let $|S|$ denote the number of elements it has.
We will show

\begin{stmt}
For any set of feature vectors $S \subseteq \{\x_1, \x_2, \ldots, \x_n\}$
with $|S| > 1$, we have 
$\avgf{f}{S_A} > \avgf{f}{S_D}$.
\label{stmt:cond-exp-d}
\end{stmt}

Note that for the case when $|S| = 1$, we must have
$\avgf{f}{S_A} = \avgf{f}{S_D}$ because in this case
$S$ consists of just a single feature $\x_i$, and by assumption
$f(\x_i,A) = f(\x_i,D)$ for all $i$.

To prove \rf{stmt:cond-exp-d}, and some of the subsequent results in
this section, there is a useful way to think about averages over sets of
rows in terms of random variables.
We define the random variable $\rv_A$ to be the $f$-value of an applicant
drawn uniformly at random from group $A$, and
the random variable $\rv_D$ to be the $f$-value of an applicant
drawn uniformly at random from group $D$.
Both of these random variables take values in the set
$\{y_1, \ldots, y_n\}$, but they have different distributions over this set;
in particular, 
$\Prb{\rv_A = y_j} = \ms(\x_j,A) / \sum_{i=1}^n \ms(\x_i,A)$ and 
$\Prb{\rv_D = y_j} = \ms(\x_j,D) / \sum_{i=1}^n \ms(\x_i,D)$.
We will use $\alpha_j$ to denote $\Prb{\rv_A = y_j}$ and
$\delta_j$ to denote $\Prb{\rv_D = y_j}$.
Note that the disadvantage condition \rf{stmt:disad} implies
that the sequence of ratios $\alpha_i / \delta_i$ is 
strictly increasing in $i$: 
if $j > i$, then $\alpha_j / \delta_j > \alpha_i / \delta_i$.

In the language of random variables, 
the disadvantage condition thus asserts that
the random variable $\rv_A$ exhibits 
{\em likelihood-ratio dominance} with respect to the random
variable $\rv_D$
\cite{athey-monotone-likelihood,hopkins-ratio-orderings,milgrom-good-news,wolfstetter-microeconomics-book}.
It is a standard fact from this literature
that if one random variable likelihood-ratio
dominates another, then it also has a strictly greater expected value
\cite{hopkins-ratio-orderings,wolfstetter-microeconomics-book}.
We record this fact here in a general form, since we will need it in 
some of the subsequent arguments.

\begin{stmt} (See e.g. \cite{hopkins-ratio-orderings,wolfstetter-microeconomics-book})
Consider two discrete random variables $\rvd$ and $\rva$,
each of which takes values in $\{\rvval_1, \rvval_2, \ldots, \rvval_n\}$,
with $\rvval_1 < \rvval_2 < \cdots < \rvval_n$ and $n > 1$.
Let $\rvdprob_i = \Prb{\rvd_i = \rvval_i}$ and $\rvaprob_i = \Prb{\rva_i = \rvval_i}$;
so $\sum_{i = 1}^n \rvdprob_i = \sum_{i = 1}^n \rvaprob_i = 1$, and
the expected values are given by
$\Exp{\rvd} = \sum_{i = 1}^n \rvdprob_i \rvval_i$ and
$\Exp{\rva} = \sum_{i = 1}^n \rvaprob_i \rvval_i$.
We will assume that $\rvdprob_i > 0$ and $\rvaprob_i > 0$ for all $i$.

If the sequence of ratios $\{\rvaprob_i/\rvdprob_i\}$ is 
strictly monotonically increasing 
then $\Exp{\rva} > \Exp{\rvd}$.
\label{stmt:rv-compare-exp}
\end{stmt}

For completeness, we give a proof \rf{stmt:rv-compare-exp}
in the appendix.
Using this fact, we can now give a proof of
\rf{stmt:cond-exp-d}.

\prevs{stmt:cond-exp-d}{
In the language of random variables, 
\rf{stmt:rv-compare-exp} is equivalent to showing that
for every set $\yset \subseteq \{y_1, \ldots, y_n\}$ with $|\yset| > 1$, 
we have $\Expg{Y_A}{Y_A \in \yset} > \Expg{Y_D}{Y_D \in \yset}$.

To prove this, we
write $\yset = \{y_{i_1}, y_{i_2}, \ldots, y_{i_c}\}$
for $i_1 < i_2 < \cdots < i_c \subseteq \{1, 2, \ldots, n\}$.
Let us define $Y_D\^{\yset}$ to be the random variable defined on $\yset$ by
$$\Prb{Y_D\^{\yset} = y_{i_j}} = \frac{\delta_{i_j}}{\sum_{\ell = 1}^c \delta_{i_{\ell}}}.$$
We define $Y_A\^{\yset}$ analogously by
$$\Prb{Y_A\^{\yset} = y_{i_j}} = \frac{\alpha_{i_j}}{\sum_{\ell = 1}^c \alpha_{i_{\ell}}}.$$
We observe that $\Exp{Y_D\^{\yset}} = \Expg{Y_D}{Y_D \in \yset}$ and
$\Exp{Y_A\^{\yset}} = \Expg{Y_A}{Y_A \in \yset}$.
Moreover, we have
$$\Prb{Y_A\^{\yset} = y_{i_j}} / \Prb{Y_D\^{\yset} = y_{i_j}} =
\frac
  {\frac{\alpha_{i_j}}{\sum_{\ell = 1}^c \alpha_{i_{\ell}}}}
  {\frac{\delta_{i_j}}{\sum_{\ell = 1}^c \delta_{i_{\ell}}}} =
\frac
  {\alpha_{i_j} \sum_{\ell = 1}^c \delta_{i_{\ell}}}
  {\delta_{i_j} \sum_{\ell = 1}^c \alpha_{i_{\ell}}},$$
where the second term in each of the numerator and denominator
is independent of $j$; thus, this sequence of ratios is strictly
monotonically increasing in $j$ because $\alpha_{i_j} / \delta_{i_j}$ is.
It follows that the likelihood ratio dominance
condition as stated in \rf{stmt:rv-compare-exp} holds for
the pair of random variables $Y_A\^{\yset}$ and $Y_D\^{\yset}$.
Hence by \rf{stmt:rv-compare-exp}, we have
$\Expg{Y_A}{Y_A \in \yset} = \Exp{Y_A\^{\yset}} > \Exp{Y_D\^{\yset}} = \Expg{Y_D}{Y_D \in \yset}.$
}

\subsection{A Combinatorial Lemma about Separable Approximators}

Recall that an $f$-approximator $g$ is called {\em separable} if each non-trivial cell consists entirely of rows associated with group $A$ or entirely of rows associated with group $D$.
As a key step in the proof of \rf{stmt:graded-improvable}, we will need
the following fact: in any non-trivial, separable $f$-approximator $g$,
there exists an $A$-applicant who
receives a value that is strictly higher than a $D$-applicant
with the same feature vector.
Given the disadvantage condition, it is 
intuitively plausible that this should be true. 
But given that a number of quite similar-sounding
statements are in fact false --- essentially, these statements
are very close to what arises in {\em Simpson's Paradox} 
\cite{blyth-simpsons-paradox} ---
some amount of care is needed in the proof of this fact.
(We explore the connection to Simpson's Paradox in 
Section \ref{subsec:disad}.)

\begin{stmt}
Let $g$ be a non-trivial, separable $f$-approximator.
Then there exists an $x_j$ such that $g$ assigns the row $(x_j,A)$ 
a strictly higher value than it assigns the row $(x_j,D)$.
That is, $(x_j,A)$ and $(x_j,D)$ belong to 
cells $C_a$ and $C_b$ respectively, and $\gfn(C_a) > \gfn(C_b)$,
where as before $\gfn(C)$ denotes the average $f$-value 
of the members of a cell $C$.
\label{stmt:separable}
\end{stmt}

\proof{
Let the cells containing rows of group $D$ be $S_1, S_2, \ldots, S_c$,
and the cells containing rows of group $A$ be $T_1, T_2, \ldots, T_m$.
We define $S(j)$ to be the cell $S_i$ for which $(x_j,D) \in S_i$, and
we define $T(j)$ to be the cell $T_\ell$ for which $(x_j,A) \in T_\ell$.
For a set of rows $M$, we also define $\cv(M) = \{f(\xb) : \xb \in M\}$.

We take care of two initial considerations at the outset.
First, $g$ may contain trivial cells of the form 
$\{(x_j, A), (x_j, D)\}$, since separability only requires
that each non-trivial cell consist entirely of rows from the same group.
We can modify $g$ so that any such trivial cell is replaced
instead by the two cells $\{(x_j, A)\}$ and $\{(x_j, D)\}$.
If we obtain the result for this modified approximator, it will
hold for the original approximator $g$ as well.
Thus, we will henceforth assume that each cell of $g$ (trivial
or non-trivial) consists entirely of rows from the same group.

Second, the result is immediate in the case when all the cells $T_\ell$
associated with group $A$ are singletons.
Indeed, in this case, since $g$ is non-trivial, there must be 
a cell $S_i$ associated with group $D$ that contains more than one row.
We choose such a cell $S_i$,
let $(x_j,D)$ be the row in $S_i$ of maximum $f$-value, and
let $T_\ell$ be the (singleton) cell consisting of $\{(x_j,A)\}$.
Then $\gfn(S_i) < f(\x_j) = \gfn(T_\ell)$, and the result follows.
Thus, we will also henceforth assume that at least one cell of $g$ contains
multiple rows of $A$.

With these two preliminaries out of the way, we proceed with the
main portion of the proof.
We again use the random-variable interpretation, 
in which $\rv_A$ is the 
$f$-value of a candidate
drawn at random from group $A$, and
$\rv_D$ is the $f$-value of a candidate
drawn at random from group $D$.
Thus we have $\gfn(S_i) = \Expg{\rv_D}{\rv_D \in \cv(S_i)}$ and 
$\gfn(T_\ell) = \Expg{\rv_A}{\rv_A \in \cv(T_\ell)}$.

The statement we are trying to prove requires that we find a choice of 
$j$ for which the cell containing $(x_j,D)$ has a strictly
lower $\gfn$-value than the row containing $(x_j,A)$ ---
that is, a $j$ such that $\gfn(S(j)) < \gfn(T(j))$.
Using the connection to random variables as just noted, 
this means we need to find a $j$ for which 
$\Expg{\rv_D}{\rv_D \in \cv(S(j))} < \Expg{\rv_A}{\rv_A \in \cv(T(j))}$.

A useful start is to write

\begin{eqnarray}
\Exp{\rv_D} & = & \sum_{i = 1}^c \Expg{\rv_D}{\rv_D \in \cv(S_i)} \Prb{\rv_D \in \cv(S_i)} \nonumber \\
 & = & \sum_{i = 1}^c \left[\Expg{\rv_D}{\rv_D \in \cv(S_i)} \sum_{(x_j,D) \in S_i} \delta_j \right] \nonumber \\
 & = & \sum_{j = 1}^n \delta_j \Expg{\rv_D}{\rv_D \in \cv(S(j))}
\label{eq:cond-exp}
\end{eqnarray}

and analogously, for $\rv_A$, we have

\begin{eqnarray}
\Exp{\rv_A} & = & \sum_{\ell = 1}^m \Expg{\rv_A}{\rv_A \in \cv(T_\ell)} \Prb{\rv_A \in \cv(T_\ell)} \nonumber \\
 & = & \sum_{\ell = 1}^m \left[\Expg{\rv_A}{\rv_A \in \cv(T_\ell)} \sum_{(x_j,A) \in T_\ell} \alpha_j \right] \nonumber \\
 & = & \sum_{j = 1}^n \alpha_j \Expg{\rv_A}{\rv_A \in \cv(T(j))}
\end{eqnarray}

Given that $\Exp{\rv_D} < \Exp{\rv_A}$, this immediately tells us
that there is a $j$ for which 
\begin{equation*}
\delta_j \Expg{\rv_D}{\rv_D \in \cv(S(j))} < 
\alpha_j \Expg{\rv_A}{\rv_A \in \cv(T(j))}.
\end{equation*}
But this doesn't actually get us very far, because the terms we care about 
($\Expg{\rv_D}{\rv_D \in \cv(S(j))}$ and $\Expg{\rv_A}{\rv_A \in \cv(T(j))}$)
are being multipled by different coefficients on the two sides of
the inequality ($\delta_j$ and $\alpha_j$ respectively).
This is a non-trivial point, since in fact the statement we are
trying to prove would not in fact hold if the only thing we knew
about the random variables $\rv_D$ and $\rv_A$ were the inequality
$\Exp{\rv_D} < \Exp{\rv_A}$.  
(We explore this point further in Section \ref{subsec:disad}.)
Thus, we must use additional structure in the values of $\delta_j$ and
$\alpha_j$; in particular, we will apply the disadvantage condition
\rf{stmt:disad} and its consequence \rf{stmt:cond-exp-d}.

The idea will be to interpose a new quantity 
that we can compare with both 
$\Expg{\rv_D}{\rv_D \in \cv(S(j))}$ and $\Expg{\rv_A}{\rv_A \in \cv(T(j))}$
for any given index $j$,
and which in this way will allow us to compare these two quantities
to each other by transitivity.
To do this, we first observe that Equation (\ref{eq:cond-exp}) applies
to any partition of the rows of group $D$.
We therefore invoke this equation for a second
partition of the rows of group $D$ --- in particular, we will partition
the rows of $D$ in a way that ``lines up'' with the
partition $T_1, T_2, \ldots, T_m$ used for the rows of group $A$.
With this in mind, 
we define the following partition of the rows associated with $D$:
we write $T_\ell' = \{(x_j,D) : (x_j,A) \in T_\ell\}$.
As above, we define $T'(j)$ to be the set $T_\ell'$ 
for which $(x_j,D) \in T_\ell'$.
Following the same argument as in 
Equation (\ref{eq:cond-exp}), we have 

$$\Exp{\rv_D} = \sum_{j = 1}^n \delta_j \Expg{\rv_D}{\rv_D \in \cv(T'(j))}.$$

Subtracting this from Equation (\ref{eq:cond-exp}) for $\Exp{\rv_D}$, we get
\begin{equation}
\sum_{j = 1}^n \delta_j \left(\Expg{\rv_D}{\rv_D \in \cv(S(j))} - \Expg{\rv_D}{\rv_D \in \cv(T'(j))} \right) = 0.
\label{eq:two-exp}
\end{equation}

It will turn out to matter in the remainder of the proof
whether or not the index $j$ we are working with
has the property that $T'(j)$ is a singleton set 
(i.e. with $|T'(j)| = 1$).
Therefore, viewing the left-hand side of Equation (\ref{eq:two-exp}) as a
sum over $n$ terms, we group these terms into two sets:
let $K$ be the sum over all terms $j$
for which $T'(j)$ is a singleton,
and let $L$ be the sum over all terms $j$ for which $|T'(j)| > 1$.
Recall that since we addressed the case in which all sets
$T_\ell$ (and hence all sets $T_\ell'$) are singletons, we can
assume that at least one of the sets $T_\ell'$ has size greater than 1.
Thus, the quantity $L$ is a sum over a non-empty set of terms.
In the event that there are no singleton sets $T_\ell'$
(in which case there are no terms contributing to the value of $K$),
we declare $K = 0$.
Now, 
the left-hand side of Equation (\ref{eq:two-exp}) by definition is $K + L$,
and so $K + L = 0$. 
Thus, we cannot have both $K \geq 0$ and $L > 0$, and so
one of $K < 0$ or $L \leq 0$ must hold.
If $K < 0$, then there must be a $j^*$ for which 
$T(j^*)$ is a singleton and 
$\Expg{\rv_D}{\rv_D \in \cv(S(j^*))} < \Expg{\rv_D}{\rv_D \in \cv(T'(j^*))}$.
Alternately, if $L \leq 0$,
then there is a $j^*$ for which $T(j^*)$ is not a singleton, and 
$\Expg{\rv_D}{\rv_D \in \cv(S(j^*))} \leq \Expg{\rv_D}{\rv_D \in \cv(T'(j^*))}$.

In summary, we have thus found an index $j^*$ for which 
\begin{equation}
\Expg{\rv_D}{\rv_D \in \cv(S(j^*))} \leq \Expg{\rv_D}{\rv_D \in \cv(T'(j^*))}.
\label{eq:sep-d}
\end{equation}
and with the additional property that the inequality is strict
in the case that $T'(j^*)$ is a singleton.

We now claim that for this $j^*$, we have 
\begin{equation}
\Expg{\rv_D}{\rv_D \in \cv(S(j^*))} < \Expg{\rv_A}{\rv_A \in \cv(T(j^*))}.
\label{eq:sep-ad}
\end{equation}
As discussed at the outset of the proof,
this will establish the result, since 
it says that $(x_{j^*},D)$ and $(x_{j^*},A)$ belong to cells 
$S(j^*)$ and $T(j^*)$ respectively, and 
$$\gfn(S(j^*)) = \Expg{\rv_D}{\rv_D \in \cv(S(j^*))} 
  < \Expg{\rv_A}{\rv_A \in \cv(T(j^*))} = \gfn(T(j^*)).$$

We establish this claim by considering two cases.

\medskip 

{\bf Case 1:} {\em $|T'(j^*)| > 1$.}
In this case we can apply 
\rf{stmt:cond-exp-d} to conclude that
$\Expg{\rv_D}{\rv_D \in \cv(T'(j^*))} < \Expg{\rv_A}{\rv_A \in \cv(T(j^*))}$.
Combining this with Inequality (\ref{eq:sep-d}), we obtain
Inequality (\ref{eq:sep-ad}) by transitivity.

\medskip 

{\bf Case 2:} {\em $|T'(j^*)| = 1$.}
Above, we noted that our choice of $j^*$ ensures that 
Inequality (\ref{eq:sep-d}) is strict when
$T'(j^*)$ is a singleton, and so we have
$$\Expg{\rv_D}{\rv_D \in \cv(S(j^*))} < \Expg{\rv_D}{\rv_D \in \cv(T'(j^*))}$$
in this case.
Since $T'(j^*)$ is a singleton, consisting only of the row $(\x_{j^*},D)$,
we also have 
$$\Expg{\rv_D}{\rv_D \in \cv(T'(j^*))} = \Expg{\rv_A}{\rv_A \in \cv(T(j^*))},$$
since both the left- and right-hand sides are equal to $f(\x_{j^*})$.
Combining this with the previous inequality, we obtain
Inequality (\ref{eq:sep-ad}) in this case as well.
} 

\subsection{Proof}

We now have all the ingredients needed for proving the 
first main result.

\prevs{stmt:graded-improvable}{
Let $g$ be a graded $f$-approximator with cells $C_1, \ldots, C_d$;
thus $g$ is discrete $f$-approximator $g$ with (by non-triviality)
at least one cell containing rows $\xb, \xb'$ such that $f(\xb) \neq f(\xb')$,
and such that for every cell $C_i$, we have
$C_i\^{A} \subseteq C_i\^{D}$
or $C_i\^{D} \subseteq C_i\^{A}$.
We will create a new $f$-approximator $g'$ that strictly improves on $g$.

For an index $\ell$, recall that $r_\ell\^{g}$ is the measure of the first
$\ell$ entries in the list of cells of $g$.
It is also useful to introduce a further piece of notation for the proof:
we write $\efn_g(r) = \int_0^r \efd_g(t) ~ dt$ for the
unnormalized version of $\efc_g(r)$ in which we do not divide by $r$,
and we write $\eqn_g(r) = \int_0^r \eqd_g(t) ~ dt$ analogously.
In order to show that an approximator $g'$ improves on $g$,
we can compare the pairs of 
functions $\efn_g, \efn_{g'}$ and $\eqn_g, \eqn_{g'}$ 
rather than $\efc_g, \efc_{g'}$ and $\eqc_g, \eqc_{g'}$ 
in the underlying definition.
That is, it would be equivalent to our earlier definitions of
improvement to say that $g'$ weakly improves on $g$ 
if $\efn_{g'}(r) \geq \efn_g(r)$ and $\eqn_{g'}(r) \geq \eqn_g(r)$
for all $r \in (0,1]$;
and $g'$ strictly improves on $g$ if 
$g'$ weakly improves on $g$, and there exists $r^* \in (0,1]$ for which
$\efn_{g'}(r^*) > \efn_g(r^*)$ and $\eqn_{g'}(r^*) > \eqn_g(r^*)$.

Inside the proof, 
it will also be useful to work with objects that are slightly
more general than $f$-approximators (although the statement
of the result itself applies only to $f$-approximators as we have
defined them thus far).
In particular, we will say that $h$ is an 
{\em $f$-pseudo-approximator} if it can be obtained from
an $f$-approximator $g$ by possibly rearranging the order of the cells
so that they are no longer in decreasing order of $\gfn$-values.
We can still consider admissions rules based on pseudo-approximators $h$
just as we have for approximators $g$: applicants are admitted
according to the sequence of cells in $h$, even though they no longer
have decreasing $\gfn$-values.
We can also still define
$\efd_h, \efn_h, \eqd_h$, and $\eqn_h$ for pseudo-approximators
just as we do for approximators,
and use them in the definitions of weak and strict improvement.

We organize the proof into a set of cases.  
Each case follows the structure outlined in the discussion
after the statement of \rf{stmt:simple-improvable}: we find
a row --- or a small portion of a row --- that we can break loose
from its current cell and convert into a cell on its own;
we then place it at the position determined by its value in 
the overall ordering of cells so as to strictly improve the effiency and
the equity of the approximator.
Depending on the structure of the initial approximator $g$,
we will go about selecting the row to use for this improvement 
in different ways. 
This distinction is what determines the
decomposition of the proof into cases, but the cases otherwise
follow a parallel structure.

\medskip

{\bf Case 1:} 
{\em There is a non-trivial
cell $C_i$ such that both $C_i\^{A}$ and $C_i\^{D}$ are non-empty, and $C_i\^{A} \subseteq C_i\^{D}$.}
Of all the rows $\xb \in C_i$, we choose one of maximum $f(\xb)$.
For such an $\xb$, we have $f(\xb) > \gfn(C_i)$, since
$\gfn(C_i)$ is an average of $f$-values from multiple rows.
Also, at least one row of maximum $f$-value must be associated
with group $D$, since
for all $(\x_j,A) \in C_i$ we also have $(\x_j,D) \in C_i$;
we choose $\xb$ so that it is associated with group $D$.

From this row $\xb = (x_j,D)$, 
we create a new (non-discrete) $f$-approximator $g'$ as follows.
For a small value $\eps > 0$, we create a new cell $C'$ that contains
an $\eps$ measure of row $\xb$ and nothing else.
We correspondingly subtract an $\eps$ measure of row $\xb$ from cell $C_i$,
creating a new cell $C_i'$.
This defines the new approximator $g'$.

These new cells have the property that $\gfn(C') > \gfn(C_i) > \gfn(C_i')$,
since $\gfn(C_i)$ is a weighted average of $f$-values among which
$\gfn(C') = f(\xb)$ is the largest.
The new cell $C'$ thus moves ahead of $C_i$ in the sorted order,
to position $s < i$.
By the genericity condition \rf{stmt:genericity},
we know that $\gfn(C_i)$ is distinct from the $\gfn$-value of
all other cells, and so 
by choosing $\eps$ sufficiently small, $C_i'$ will retain its
position in the sorted order of the other cells.

The new approximator has cells 
$$C_1, \ldots, C_{s-1}, C', C_s, 
\ldots, C_{i-1}, C_i', C_{i+1}, \ldots, C_d$$
in sorted order.
Observe that $r_{s-1}\^{g}$
is the measure
of the cells in the list preceding $C'$, and
$r_{i}\^{g}$ is the measure of the cells through $C_i'$.
(For this latter point, note that there are $i+1$ entries in the list
of cells of $g'$ through $C_i'$, but since two of these cells are
a partition of $C_i$, the total measure of these $i+1$ cells is $r_{i}\^{g}$.)

For comparing the functions $\efn_g$ and $\efn_{g'}$, it
is useful to interpose the following pseudo-approximator $h$.
The pseudo-approximator $h$ 
is obtained by writing the cells of $g'$ in the order
$$C_1, \ldots, C_{s-1}, C_s, \ldots, C_{i-1}, C', C_i', C_{i+1}, \ldots, C_d$$
rather than in their sorted order.
As noted above, we can 
still define $\efd_h, \efn_h, \eqd_h$, and $\eqn_h$ as before,
and use them in the definition of weak and strict improvement.

We observe first that $\efn_g$ and $\efn_h$ agree outside the
interval $[r_{i-i}\^{g},r_i\^{g}]$, and inside this interval
we have $\efn_h(r) > \efn_g(r)$, since $\gfn(C') > \gfn(C_i')$.
Similarly, $\eqn_g$ and $\eqd_h$ agree outside the
interval $[r_{i-i}\^{g},r_i\^{g}]$, and inside this interval
we have $\eqn_h(r) > \eqn_g(r)$, since $\sb(C') = 1$ and
$\sb(C_i') < 1$.
Thus, $h \succ g$.

Now, we obtain $g'$ from $h$ by moving the cell $C'$ forward 
to position $s$ in the sorted order.
The functions $\efn_{g'}$ and $\efn_h$ are thus the same for 
$r \leq r_{s-1}\^{g}$ and $r \geq r_{i}\^{g}$.
In the interval $(r_{s-1}\^{g}, r_i\^{g})$, 
they differ in that we have moved the cell $C'$ forward to the
beginning of this interval.
For all $r \in (r_{s-1}\^{g}, r_i\^{g})$, we have
$\gfn(C') \geq \gfn(C_{j(r)})$ because $C'$ comes earlier in the sorted order,
and $1 = \sb(C') \geq \sb(C_{j(r)})$.
Thus, $\efn_{g'}(r) \geq \efn_h(r)$ and $\eqn_{g'}(r) \geq \eqn_h(r)$
for all $r \in (r_{s-1}\^{g}, r_i\^{g})$ and so we have
$g' \succeq h$.

By our result \rf{stmt:improve-transitive} on transitivity, it follows that $g' \succ g$.

\medskip

{\bf Case 2:} 
{\em There is a cell $C_i$ such that both $C_i\^{A}$ and $C_i\^{D}$ are non-empty, and $C_i\^{D} \subseteq C_i\^{A}$.}
We proceed by close analogy with Case 1, except that instead of
removing a small measure of a row $(x_j,D)$ of high $f$-value from $C_i$,
we remove a small measure of a row $(x_j,A)$ of low $f$-value.
Specifically, 
of all the rows $\xb \in C_i$, we choose one of minimum $f(\xb)$.
For this $\xb$, we have $f(\xb) < \gfn(C_i)$, and we can choose
$\xb$ to have the form $(x_j,A)$, since 
for all $(\x_j,D) \in C_i$ we also have $(\x_j,A) \in C_i$.

For a small value $\eps > 0$, we create a new cell $C'$ that contains
an $\eps$ measure of row $\xb$ and nothing else.
We correspondingly subtract an $\eps$ measure of row $\xb$ from cell $C_i$,
creating a new cell $C_i'$.
This defines the new approximator $g'$.
The new cell $C'$ moves after $C_i$ in the sorted order,
to position $t > i$, 
and by choosing $\eps$ sufficiently small, $C_i'$  retains its
position; so the new approximator $g'$ has cells 
$$C_1, \ldots, C_{i-1}, C_i', C_{i+1}, \ldots, C_{t-1}, C', C_t, \ldots, C_d.$$

As in Case 1, we interpose a pseudo-approximator $h$ with the same cells as
$g'$, but in an order that is not necessarily sorted:
$$C_1, \ldots, C_{i-1}, C_i', C', C_{i+1}, \ldots, C_{t-1}, C_t, \ldots, C_d.$$
Now, $\efn_g$ and $\efn_h$ agree outside the
interval $[r_{i-i}\^{g},r_i\^{g}]$, and inside this interval
we have $\efn_h(r) > \efn_g(r)$, since $\gfn(C_i') > \gfn(C')$.
Similarly, $\eqn_g$ and $\eqn_h$ agree outside the
interval $[r_{i-i}\^{g},r_i\^{g}]$, and inside this interval
we have $\eqn_h(r) > \eqn_g(r)$, since $\sb(C_i') > 0$ and
$\sb(C') = 0$.
Thus, $h \succ g$.

Now, we obtain $g'$ from $h$ by moving the cell $C'$ back
to position $t$ in the sorted order.
The functions $\efn_{g'}$ and $\efn_h$ are thus the same for 
$r \leq r_{i-1}\^{g}$ and $r \geq r_{t-1}\^{g}$.
In the interval $(r_{i-1}\^{g}, r_{t-1}\^{g})$, 
they differ in that we have moved the cell $C'$ back to the
end of this interval.
For all $r \in (r_{i-1}\^{g}, r_{t-1}\^{g})$, we have
$\gfn(C') \leq \gfn(C_{j(r)})$ because $C'$ comes later in the sorted order,
and $0 = \sb(C') < \sb(C_i')$.
Thus, $\efn_{g'}(r) \geq \efn_h(r)$ and $\eqn_{g'}(r) \geq \eqn_h(r)$
for all $r \in (r_{i-1}\^{g}, r_{t-1}\^{g})$ and so we have
$g' \succeq h$.  By transitivity, it follows that $g' \succ g$.

\medskip

If neither of Cases 1 or 2 holds, then every non-trivial cell of
$g$ consists entirely of rows associated with $A$ or entirely
of rows associated with $D$.
Thus our final case is the following.

\medskip

{\bf Case 3:} 
{\em $g$ is separable.}
In this case, \rf{stmt:separable} implies that there is an $x_j$
such that $(x_j,A)$ and $(x_j,D)$ belong to cells
$C_a$ and $C_b$ respectively, and $\gfn(C_a) > \gfn(C_b)$.
This case follows a similar high-level strategy to
Cases 1 and 2, but using the existence of $x_j$ and the cells
$C_a$ and $C_b$ to create a new cell.  The construction of this new
cell involves moving a small amount
of measure out of $C_a$ or $C_b$, and in some cases out of both cells.
For this case, we consider a set of sub-cases, depending on the relative
ordering of $f(x_j)$ with respect to $\gfn(C_a)$ and $\gfn(C_b)$.

\medskip

{\bf Case 3a:} {\em $f(x_j) \geq \gfn(C_a) > \gfn(C_b)$.}
We define an $f$-approximator $g'$ by creating a new cell $C'$ that,
for a small $\eps > 0$,
contains an $\eps$ measure of row $(x_j,D)$ and nothing else.
We correspondingly subtract an $\eps$ measure of row $(x_j,D)$ from cell $C_b$,
creating a new cell $C_b'$.
We put $C'$ in its appropriate place in the sorted order of cells, 
and ahead of $C_a$ in the case that $f(x_j) = \gfn(C_a)$;
so $C'$ goes into some position $s \leq a$.
The cells of $g'$ in order are thus
$$C_1, \ldots, C_{s-1}, C', C_{s}, \ldots, C_a, \ldots, C_{b-1}, C_b', C_{b+1}, \ldots, C_d,$$
where possibly $C_s = C_a$.
We also consider the pseudo-approximator $h$ with these same cells in the order
$$C_1, \ldots, C_{s-1}, C_{s}, \ldots, C_a, \ldots, C_{b-1}, C', C_b', C_{b+1}, \ldots, C_d.$$
The functions $\efn_{g}$ and $\efn_{h}$ agree outside the interval 
$(r_{b-1}\^g, r_b\^g)$, and for $r$ in this interval we have 
$\efn_h(r) > \efn_g(r)$.
When we then move $C'$ forward to its correct position in sorted order, 
producing $g'$, we see that
the functions $\efn_{g}$ and $\efn_{g'}$ agree outside the interval
$(r_{s-1}\^{g},r_b\^{g})$; and for $r$ in this interval, we have 
$\efn_{g'}(r) > \efn_{g}(r)$.

Similarly, the functions $\eqn_{g}$ and $\eqn_{g'}$ agree for 
$r \leq r_{s-1}\^{g}$ and $r \geq r_b\^{g}$.
For $r_{s-1}\^{g} < r < r_b\^{g}$, the functions $\eqn_{g}$ and $\eqn_{g'}$
differ in the fact that cell $C'$ with $\sb(C') = 1$ was moved ahead of $C_s$,
so for all such $r$ we have $\eqn_{g'}(r) \geq \eqn_{g}(r)$.
Now, let $C_\ell$ be the next cell after $C'$ for which $\sb(C_\ell) < 1$;
we have $\ell \leq a$, since $\sb(C_a) = 0$.
For $r_{\ell}\^{g'} < r < r_{\ell+1}\^{g'}$, the marginal applicant
is being drawn from $C_\ell$, and hence 
$\eqn_{g'}(r) > \eqn_{g}(r)$ for such $r$.
Since this $r$ is also in the range noted above where
$\efn_{g'}(r) > \efn_{g}(r)$,
we have $g' \succ g$.

\medskip

{\bf Case 3b:} {\em $\gfn(C_a) > \gfn(C_b) \geq f(x_j)$.}
We define an $f$-approximator $g'$ by creating a new cell $C'$ that,
for a small $\eps > 0$,
contains an $\eps$ measure of row $(x_j,A)$ and nothing else.
We correspondingly subtract an $\eps$ measure of row $(x_j,A)$ from cell $C_a$,
creating a new cell $C_a'$.
We put $C'$ in its appropriate place in the sorted order of cells, 
and after $C_b$ in the case that $f(x_j) = \gfn(C_b)$;
so $C'$ goes into some position $t > b$.
The cells of $g'$ in order are thus
$$C_1, \ldots, C_{a-1}, C_a', C_{a+1}, \ldots, C_b, \ldots, C_{t-1}, C', C_{t}, \ldots, C_d.$$
We also consider the pseudo-approximator $h$ with these same cells in the order
$$C_1, \ldots, C_{a-1}, C_a', C', C_{a+1}, \ldots, C_b, \ldots, C_{t-1}, C_{t}, \ldots, C_d.$$
The functions $\efn_{g}$ and $\efn_{h}$ agree outside the interval 
$(r_{a-1}\^g, r_a\^g)$, and for $r$ in this interval we have 
$\efn_h(r) > \efn_g(r)$.
When we then move $C'$ back to its correct position in sorted order, 
producing $g'$, we see that
the functions $\efn_{g}$ and $\efn_{g'}$ agree outside the interval
$(r_{a-1}\^{g},r_{t-1}\^{g})$; and for $r$ in this interval, we have 
$\efn_{g'}(r) > \efn_{g}(r)$.

The functions $\eqn_{g}$ and $\eqn_{g'}$ agree for $r \leq r_{a-1}\^{g}$
and $r \geq r_{t-1}\^{g}$.
For $r_{a-1}\^{g} < r < r_{t-1}\^{g}$, 
the functions $\eqn_{g}$ and $\eqn_{g'}$
differ in the fact that cell $C'$ with $\sb(C') = 0$ was moved after $C_{t-1}$,
so for all such $r$ we have $\eqn_{g'}(r) \geq \eqn_{g}(r)$.
Now, let $C_\ell$ be the next cell after $C_a'$ for which $\sb(C_\ell) > 0$;
we have $\ell \leq b$, since $\sb(C_b) = 1$.
For $r_{\ell}\^{g'} < r < r_{\ell+1}\^{g'}$, the marginal applicant
is being drawn from $C_\ell$, and hence 
$\eqn_{g'}(r) > \eqn_{g}(r)$ for such $r$.
Since this $r$ is also in the range noted above where
$\efn_{g'}(r) > \efn_{g}(r)$,
we have $g' \succ g$.

\medskip

{\bf Case 3c:} {\em $\gfn(C_a) > f(x_j) > \gfn(C_b).$}
In this case, 
we define an $f$-approximator $g'$ by creating a new cell $C'$ that,
for a small $\eps > 0$,
contains an $\eps$ measure of row $(x_j,A)$ and an 
$\eps$ measure of row $(x_j,D)$.
We correspondingly subtract an $\eps$ measure of row $(x_j,A)$ from cell $C_a$,
creating a new cell $C_a'$, and we subtract
an $\eps$ measure of row $(x_j,D)$ from cell $C_b$,
creating a new cell $C_b'$.
We put $C'$ in its appropriate place in the sorted order of cells, 
which is in some position $i$ with $a < i \leq b$.
The cells of $g'$ in order are thus
$$C_1, \ldots, C_{a-1}, C_a', \ldots, C', \ldots, C_b', C_{b+1}, \ldots, C_d.$$

We consider two pseudo-approximators $h$ and $h'$.
For these, we define $C_a^+$ to be a cell consisting only of an 
$\eps$ measure of row $(x_j,A)$, and we define
$C_b^+$ to be a cell consisting only of an 
$\eps$ measure of row $(x_j,D)$.
Note that $C'$ is obtained by merging $C_a^+$ and $C_b^+$ together.
We define $h$ to have the sequence of cells
$$C_1, \ldots, C_{a-1}, C_a', C_a^+, \ldots, C_b^+, C_b', C_{b+1}, \ldots, C_d.$$
We define $h'$ to be obtained from $h$ by shifting $C_a^+$ later
if necessary and $C_b^+$ forward if necessary so they are 
consecutive in position $i$.

Now, the functions $\efn_g$ and $\efn_h$ agree outside the intervals
$(r_{a-1}\^{g},r_a\^{g})$ and $(r_{b-1}\^{g},r_b\^{g})$;
inside these intervals we have $\efn_h(r) > \efn_g(r)$.
When we move $C_a^+$ and $C_b^+$ to be adjacent in position $i$,
we obtain $h'$ with $\efn_{h'}(r) > \efn_g(r)$ for 
$r \in (r_{a-1}\^{g},r_b\^{g})$
(and the same function outside this interval).
Finally, $\efn_{h'}$ and $\efn_{g'}$ are the same function everywhere.

The functions $\eqn_{g}$ and $\eqn_{h'}$ agree for $r \leq r_{a-1}\^{g}$
and $r \geq r_{b}\^{g}$.
For $r_{a-1}\^{g} < r < r_{b}\^{g}$, 
the functions $\eqn_{g}$ and $\eqn_{h'}$
differ in the fact that cell $C_a^+$ with $\sb(C_a^+) = 0$ 
was moved after $C_{i-1}$,
and cell $C_b^+$ with $\sb(C_b^+) = 1$ was moved forward to be just behind it;
so for all $r \in (r_{a-1}\^{g}, r_{b}\^{g})$
we have $\eqn_{h'}(r) \geq \eqn_{g}(r)$.
Since $g'$ is obtained from $h'$ by simply combining the adjacent cells
$C_a^+$ and $C_b^+$ into the single cell $C'$, 
we have $\eqn_{g'}(r) \geq \eqn_{h'}(r)$ for all $r$.
Now, let $C_\ell$ be the next cell of $g'$ after $C_a'$ 
for which $\sb(C_\ell) > 0$;
$C_\ell$ must be before or equal to $C'$, since $\sb(C') = 1/2$.
For $r_{\ell-1}\^{g'} < r < r_{\ell}\^{g'}$, the marginal applicant
is from $C_\ell$, so 
$\eqn_{g'}(r) > \eqn_{g}(r)$ for such $r$.
Since this $r$ is in the range noted above where
$\efn_{g'}(r) > \efn_{g}(r)$,
we have $g' \succ g$.
}

\subsection{The Role of Non-Discrete Approximators}

The proof of \rf{stmt:graded-improvable} showed how an arbitrary
graded $f$-approximator $g$ could be improved by another 
$f$-approximator $g'$.
The construction in the proof produced $f$-approximators that
were not necessarily discrete, and it is natural to ask whether
the use of non-discrete approximators was esssential for the result;
is it possible that for every graded $f$-approximator, there
is a {\em discrete} $f$-approximator that strictly improves it?

In fact, there exist graded $f$-approximators $g$ such that every approximator
strictly improving $g$ is non-discrete; this establishes that 
non-discrete approximators are indeed necessary for the result.
In this section, we give an example of such a graded approximator.

We use the example from Figure \ref{fig:example}
in Section \ref{sec:examples}, 
with the parameters $y_{10}$ and $y_{01}$ close enough to 0
so that the following holds:
if $S$ is the set of three rows
$\{(1,1,D), (0,1,D), (0,0,D)\}$, then 
$\avgf{f}{S} > f(1,0,D)$.
For the function $f$ given in Figure \ref{fig:example}, 
consider the following $f$-approximator $g$:
it consists of cells $C_1, C_2, \ldots, C_6$, where
$C_1$ is the row $(1,1,A)$, $C_2 = S$,
and $C_3, C_4, C_5, C_6$
are each singleton sets consisting of the rows 
$(1,0,D)$, $(1,0,A)$, $(0,1,A)$, and $(0,0,A)$ respectively.

If we follow the construction used in the proof of \rf{stmt:graded-improvable},
we arrive at the following non-discrete $f$-approximator $g'$ that strictly
improves $g$. Let $C_0(\eps)$ be the cell consisting of an
$\eps$ measure of the row $(1,1,D)$, and let $C_2'(\eps)$
be the cell we obtain by starting with $C_2$ and removing an $\eps$
measure of the row $(1,1,D)$.  
We also define 
the cell $C_2''(\eps)$ to consist of $C_2'(\eps)$ together with $C_3$.
As we increase $\eps$, the value of $\gfn(C_2'(\eps))$ decreases
monotonically; and by the time $\eps$ reaches $\ms(1,1,D)$, 
we have $\gfn(C_2'(\eps)) < \gfn(C_3)$, since
for this value of $\eps$, the cell $C_2'(\eps)$ consists of just
the rows $(0,1,D)$ and $(0,0,D)$.
We can therefore find an $\eps^* > 0$ such that
$\gfn(C_2'(\eps)) = \gfn(C_3)$, and we define
the $f$-approximator $g'$ to
consist of the cells
$$C_0(\eps^*), C_1, C_2''(\eps^*), C_4, C_5, C_6.$$
$g'$ is a non-discrete $f$-approximator that strictly improves $g$.

However, it turns out that the only $f$-approximators that
strictly improve $g$ are non-discrete, as we establish in the following.

\begin{stmt}
There is no discrete $f$-approximator that strictly improves $g$.
\label{stmt:non-discrete}
\end{stmt}

\proof{
Let $h$ be a discrete $f$-approximator that weakly improves $g$.
We will show that $h$ does not strictly improve $g$.

Let $s_1 = \ms(1,1,A)$ and $s_2 = \ms(1,1,A) + 1/2$; we recall
that the measure of all rows associated with group $D$ is $1/2$.
We observe that (i) $\efc_h(s_1) \geq \efc_g(s_1) = 1$, and 
(ii) $\eqc_h(s_2) \geq \eqc_g(s_2) = 1 / (1 + 2s_1)$.
Since $\ms(1,1,A) > \ms(1,1,D)$, fact (i) implies that the row
$(1,1,A)$ must be in a cell $C$ by itself or with just $(1,1,D)$.
Fact (ii) implies that all rows associated with $D$ must occur
in cells that come before every row associated with $A$
except for $(1,1,A)$.
From this we can conclude that the row $(1,1,D)$ must occur in
a cell $C'$ together with rows $(0,1,D)$ and $(0,0,D)$;
and hence $C$ consists of just the row $(1,1,A)$.

Now, recall that $S$ is the set of rows
$\{(1,1,D), (0,1,D), (0,0,D)\}$.
Since $h$ weakly improves $g$, and since
$\avgf{f}{S} > f(1,0,D)$, it follows that $C' = S$, and the
first two cells of $h$ are $C$ and $C'$.
The remaining cells of $g$ are singleton rows, and so 
there is no value of $r$ for which $\efc_h(r) > \efc_g(r)$.
Hence $h$ does not strictly improve $g$, as required.
}

\section{Proof of Second Result: \\ Simplicity Can Transform Disadvantage Into Bias}
\label{sec:transform}

In this section we prove our second main result,
\rf{stmt:group-split}.

Since \rf{stmt:group-split} is concerned with the process of 
taking a group-agnostic $f$-approximator $g$ and splitting
its non-trivial cells by group to produce $\gpdep(g)$, it is useful
to first consider the effect of splitting a {\em single} non-trivial cell 
in this way.
By considering such single steps first, we can then analyze 
a sequence of such steps to go from $g$ to $\gpdep(g)$.

Given this plan, it is useful to introduce some terminology and notation
for individual cells.
Given a discrete $f$-approximator $g$, we say that one of its 
cells $C$ is {\em group-agnostic} if 
$C\^{A} = C\^{D}$ in the notation of the previous sections;
that is, if for all feature vectors $x$, we have 
$(x,A) \in C$ if and only if $(x,D) \in C$.
We will use 
$\gpdep_A(C)$ to denote $\{(x,A) : (x,A) \in C\}$,
and 
$\gpdep_D(C)$ to denote $\{(x,D) : (x,D) \in C\}$,
and we say $g'$ is obtained from $g$ by {\em splitting $C$} if
$g'$ has the same cells as $g$, but with $C$ replaced by the two cells
$\gpdep_A(C)$ and $\gpdep_D(C)$.

Using the disadvantage condition \rf{stmt:disad} and its
consequence \rf{stmt:cond-exp-d}, we now prove

\begin{stmt}
Let $g$ be a discrete $f$-approximator, and let $C$ be a 
non-trivial group-agnostic cell of $g$.
Let $g'$ be the approximator obtained from $g$ by splitting $C$.
Then $g'$ strictly improves $g$ in efficiency, and $g$ strictly
improves $g'$ in equity.
\label{stmt:one-split}
\end{stmt}

\proof{
Let the cells of $g$ be $C_1, \ldots, C_d$, with 
$C = C_i$ the non-trivial group-agnostic cell that we
split to obtain $g'$.
Since $C$ is non-trivial, it contains at least two rows associated
with each of groups $A$ and $D$, and hence
\rf{stmt:cond-exp-d} implies that 
$\gfn(\gpdep_A(C)) > \gfn(\gpdep_D(C))$.
Moreover, since $\gfn(C)$ is a weighted average of
$\gfn(\gpdep_A(C))$ and $\gfn(\gpdep_D(C))$, we can extend this
inequality to say
$\gfn(\gpdep_A(C)) > \gfn(C) > \gfn(\gpdep_D(C))$.

Now, $g'$ consists of the cells $C_1, \ldots, C_{i-1}, C_{i+1}, \ldots, C_d$
together with $\gpdep_A(C)$ and $\gpdep_D(C)$.
In the ordering of these cells by $\gfn$-value,
suppose that $\gpdep_A(C)$ comes just after cell $C_a$,
and $\gpdep_D(C)$ comes just before cell $C_b$, where $a < b$.
Thus, the cells of $g'$ are 
$$C_1, \ldots, C_a, \gpdep_A(C), \ldots, \gpdep_D(C), C_b, \ldots, C_d.$$
We also recall the notion of {\em pseudo-approximators} from the
proof of \rf{stmt:graded-improvable} in the previous section;
these are simply the analogues of approximators in which the cells
do not need to be arranged in descending order of $\gfn$-values.
In particular, it will be useful to consider 
the pseudo-approximator $h$ in which $\gpdep_A(C)$ and $\gpdep_D(C)$
are adjacent between $C_{i-1}$ and $C_{i+1}$; that is, $h$ has cells
$$C_1, \ldots, C_a, \ldots, C_{i-1}, 
\gpdep_A(C), \gpdep_D(C), C_{i+1}, \ldots, C_b, \ldots, C_d,$$
where either or both of $a = i-1$ or $i+1 = b$ might hold.

Now, the functions $\efn_g(r)$ and $\efn_h(r)$ agree for
$r$ outside the interval $[r_{i-1}\^{g}, r_i\^{g}]$,
and inside this interval we have $\efn_h(r) > \efn_g(r)$,
since $\gfn(\gpdep_A(C)) > \gfn(\gpdep_D(C))$.
Similarly, the functions $\eqn_g(r)$ and $\eqn_h(r)$ agree for
$r$ outside the interval $[r_{i-1}\^{g}, r_i\^{g}]$,
and inside this interval we have $\eqn_h(r) < \eqn_g(r)$,
since $\sb(\gpdep_A(C)) = 0$ and $\sb(\gpdep_D(C)) = 1$.
Thus $h$ strictly improves $g$ in efficiency, and 
$g$ strictly improves $h$ in equity.
Using our notation from earlier, we can write this as 
$h \succ_\efd g$ and $g \succ_\eqd h$.

Next, we obtain $g'$ from $h$ by moving the cell $\gpdep_A(C)$
forward so that it follows $C_a$, and moving the cell $\gpdep_D(C)$
backward to that it comes before $C_b$.
(In each case, the cell might not actually change position if
$a = i-1$ or $i+1 = b$ respectively.)
Thus, $\efn_{g'}$ and $\efn_h$ agree outside the interval
$[r_a\^{g},r_{b-1}\^{g}]$, and inside this interval we have
$\efn_{g'}(r) \geq \efn_h(r)$.
Similarly, $\eqn_{g'}$ and $\eqn_h$ agree outside the interval
$[r_a\^{g},r_{b-1}\^{g}]$, and 
since $\sb(\gpdep_A(C)) = 0$ and $\sb(\gpdep_D(C)) = 1$,
we have $\eqn_{g'}(r) \leq \efn_h(r)$ inside this interval.

It follows that $\efn_{g'}(r) \geq \efn_h(r)$ and
$\eqn_{g'}(r) \leq \efn_h(r)$ for all $r \in (0,1]$.
Since we established above that 
$h \succ_\efd g$ and $g \succ_\eqd h$, it follows 
by transitivity that
$g' \succ_\efd g$ and $g \succ_\eqd g'$.
}

We will now apply \rf{stmt:one-split} iteratively in a construction
that converts $g$ into $\gpdep(g)$ one split at a time to
prove \rf{stmt:group-split}.

\prevs{stmt:group-split}{
Let $g$ be a non-trivial, group-agnostic $f$-approximator, with cells
$C_1, C_2, \ldots, C_d$ sorted so that the values $\gfn(C_i)$
are in descending order.
Since $g$ is non-trivial, it contains at least one non-trivial cell.

We are going to imagine modifying $g$ into $\gpdep(g)$ by 
splitting one non-trivial cell at a time, 
comparing the efficiency and equity
after each individual cell is split via \rf{stmt:one-split},
and then using transitivity
to compare the efficiency at the beginning and end of the process.
As before, for a non-trivial cell $C_i$,
we let $\gpdep_A(C_i) = \{(\x,A) : (\x,A) \in C_i\}$, and 
$\gpdep_D(C_i) = \{(\x,D) : (\x,D) \in C_i\}$.
Let $g_j'$ be the $f$-approximator obtained by applying the
splitting operation to all non-trivial cells $C_i$ with $i \leq j$;
that is, we construct $g_j'$ by replacing each non-trivial cell $C_i$,
for $i \leq j$, with the two cells $\gpdep_A(C_i)$ and $\gpdep_D(C_i)$.
(For notational consistency, we will sometimes use $g_0'$ to denote $g$.)

Note that we do not need to consider the effect of splitting
any of the trivial cells of $g$: the definition of $\gpdep(g)$
involves merging together cells of the same $\gfn$-value,
and this would be true of two singleton cells of the form
$\{(\x, A)\}$ and $\{(\x, D)\}$; they would be merged together
into a single cell.
Thus, if the two-element set $\{(\x, A), (\x, D)\}$ is a cell of $g$,
then it will also be a cell of $\gpdep(g)$.
It follows that after we have split all the non-trivial cells,
we have produced $\gpdep(g)$; that is, $g_d' = \gpdep(g)$.

Now, if $C_j$ is a non-trivial cell, then $g_j'$ is obtained from
$g_{j-1}'$ by splitting the single non-trivial cell $C_j$.
If follows from \rf{stmt:one-split} that $g_j'$ strictly improves 
$g_{j-1}'$ in efficiency, and $g_{j-1}'$ strictly improves $g_j'$ in equity.
Since $g$ has at least one non-trivial cell, 
it then follows by transitivity that
$g_d' = \gpdep(g)$ strictly improves $g_0' = g$ in efficiency, and 
$g_0' = g$ strictly improves $g_d' = \gpdep(g)$ in equity,
as required.
}

\section{The Role of the Disadvantage Condition}
\label{subsec:disad}

A further natural question to ask is how much we can weaken
the disadvantage condition \rf{stmt:disad} and still derive
the conclusions of our main results.
That is, how extensive does the disadvantage of group $D$ relative
to group $A$ have to be in order for every graded approximator 
to be strictly improvable; and in order for every 
non-trivial group-agnostic approximator to create an incentive
for using group membership in a way that's biased against group $D$?

While we do not know the exact answer to this question, we can show that
it is not sufficient to require only a difference in means ---
i.e., to require only that the average $f$-value
of the $A$-applicants exceeds the average $f$-value of the $D$-applicants.
Thus, the ``boundary'' between those disadvantage conditions that
yield the main results and those that do not lies somewhere between
condition \rf{stmt:disad} and a simple difference in means.

To write this using more compact notation, let $U_A$ be the set of all
rows associated with group $A$, and $U_D$ be the set of all rows
associated with group $D$. 
We will show that there are instances
for which $\avgf{f}{U_A} > \avgf{f}{U_D}$, but for which 
(i) there exist graded approximators that are not strictly improvable, and
(ii) there are non-trivial group-agnostic approximators for which
the addition of the group membership variable benefits group $D$
rather than group $A$.
We will see, however, that the instances with this property also
reveal some of the subtleties 
inherent in defining what we mean intuitively
by {\em disadvantage.}

\begin{figure}[t]
\begin{center}
\begin{tabular}{@{}|p{0.25in}|p{0.25in}|p{0.25in}|p{0.25in}|p{0.5in}|}
\hline
$x\^{1}$ & $x\^{2}$ & $\gp$ & $f$ & $\ms$
\\\hline
1 & 1 & $D$ & .9 & .06
\\\hline
1 & 1 & $A$ & .9 & .04
\\\hline
1 & 0 & $D$ & .6 & .02
\\\hline
1 & 0 & $A$ & .6 & .06
\\\hline
0 & 1 & $D$ & .2 & .07
\\\hline
0 & 1 & $A$ & .2 & .06
\\\hline
0 & 0 & $D$ & .02 & .35
\\\hline
0 & 0 & $A$ & .02 & .34
\\\hline
\end{tabular}
\caption{An example of a function $f$ in which the average $f$-value of the
$A$-applicants exceeds the average $f$-value of the $D$-applicants;
but the simple $f$-approximator that projects out the variable $x\^{2}$
is not strictly improvable.
\label{fig:simpson}
}
\end{center}
\end{figure}

\xhdr{An Example with a Difference in Means}
The example we consider, shown in Figure \ref{fig:simpson}, is derived from the following high-level considerations.
We construct a function $f$ that depends on the two variables
$\x\^{1}$ and $\x\^{2}$ (which appear, as usual, together
with a group membership variable $\gp$ that doesn't affect 
the value of $f$).  We define $f$ so that its value
increases when either of $\x\^{1}$ or
$\x\^{2}$ is changed from $0$ to $1$; but 
the variable $x\^{1}$ has, informally,
a more significant effect on the value of $f$ than the variable $\x\^{2}$ does.
Group $A$ has more applicants with $\x\^{1} = 1$ than group $D$ does;
{\em however}, for each fixed value of $\x\^{1}$, group $D$ has
a larger fraction of applicants with $\x\^{2} = 1$.
Because we can arrange the construction so that it has these properties, the fact that $\x\^{1}$ has a larger effect on $f$ means that group $A$ will have a higher average $f$-value; but a simple $f$-approximator that ignores the value of $\x\^{2}$ will be favorable for group $D$, since it will be averaging over possible values of $\x\^{2}$ given the value of $\x\^{1}$.

Figure \ref{fig:simpson} shows the details of how we carry 
this out.\footnote{The example used in the figure does not 
satisfy the genericity condition as written --- for example,
there are two distinct subsets of rows with average value equal to $.22$ ---
but by perturbing all values very slightly we can obtain the same
relative ordering of all values and hence the same conclusions
while ensuring the genericity condition.
For ease of exposition, we explain the example using
the simpler numbers depicted in the figure.}
With numbers as in the figure, 
simple calculations show that the average $f$-value of the $A$-applicants
is $\avgf{f}{U_A} = .1816$, while the average $f$-value of the
$D$-applicants is $\avgf{f}{U_D} = .174$.
However, consider the four cells obtained by ignoring the value
of $x\^{2}$ for each applicant.
That is, for $a \in \{0,1\}$ and $b \in \{A,D\}$, we define
the cell $C_{a,b}$ to consist of all rows in which the value of
$\x\^{1}$ is $a$ and the value of the group membership variable $\gp$ is $b$.
Again, simple calculations show that 
$\gfn(C_{1,D}) = .825$;
$\gfn(C_{1,A}) = .72$;
$\gfn(C_{0,D}) = .05$; 
$\gfn(C_{0,A}) = .047$; 
and this is the ordering of the cells by $\gfn$-value.

Now, consider the $f$-approximator $g$ that uses these four cells.
From the calculations above, we can easily verify the following two claims.
\begin{itemize}
\item[(i)] $g$ is simple, and it is not strictly improvable,
since the only ways of strictly increasing its equity at any value of $r$
would have the effect of reducing its efficiency.
\item[(ii)] If we let $h$ be the $f$-approximator that uses just
the two cells $C_{1,D} \cup C_{1,A}$ and $C_{0,D} \cup C_{0,A}$,
then $h$ is non-trivial and group-agnostic, and $g = \gpdep(h)$.
But $g$ strictly improves $h$ in both efficiency and equity,
which by \rf{stmt:group-split}
cannot happen in the presence of our stronger disadvantage
condition \rf{stmt:disad}.
\end{itemize}

Intuitively, what's happened in this example is that in the
absence of any information about an applicant, our estimate
of the $f$-value of a random $A$-applicant exceeds our
estimate for a random $D$-applicant.
However, if we condition on either $\x\^{1} = 0$ or on $\x\^{1} = 1$,
we expect the random $D$-applicant to have a higher $f$-value,
since they are more likely to have $\x\^{2} = 1$.
That is, whatever we learn about the value of $\x\^{1}$, it causes
us to start favoring the random $D$-applicant over the random $A$-applicant.
The result is that each cell of $g$ consisting of rows of $D$ comes ahead
of the corresponding cell consisting of rows of $A$; this is why
$g$ is not strictly improvable, and why $g$ strictly improves $h$
in both efficiency and equity.

The surprising property that makes this example work is an
instance of {\em Simpson's Paradox} \cite{blyth-simpsons-paradox},
noted earlier, 
which roughly speaking
describes situations in which the unconditional mean over
one population $A$ exceeds the unconditional mean over another population $D$;
but there is a variable $\x$ such that when we condition on any
possible value of $\x$, the conditional mean over $D$ exceeds the
conditional mean over $A$.
It is informative to contrast this with the situation when
we impose the stronger condition \rf{stmt:disad}.
In this case, we can think of our combinatorial lemma 
\rf{stmt:separable} as showing that the structure of Simpson's
Paradox cannot happen in the presence of \rf{stmt:disad}.

The example in Figure \ref{fig:simpson} also raises the question
of when we should view a weakening of \rf{stmt:disad} as 
corresponding intuitively to disadvantage.
Specifically, although the average $f$-value of $D$-applicants in our example
is lower than the average $f$-value of $A$-applicants --- and this
is clearly a kind of disadvantage --- the example also has the 
property that the highest values of $f$, when $(\x\^{1},\x\^{2}) = (1,1)$,
are in fact characterized by an overrepresentation of group $D$.  (And more generally, group $D$ is favored once we fix either choice of value for $\x\^{1}$.)
Condition \rf{stmt:disad}, on the other hand, ensures that 
the decreasing representation of group $D$ continues at all
levels as we increase the value of $f$.
As suggested at the outset, finding the weakest version of the
disadvantage condition for which our result holds is an interesting
open question.

\section{Further Direction: Approximators with a Bounded Number of Variables}

Our result \rf{stmt:group-split} on group-agnostic approximators
showed that in the presence of disadvantage, 
an approximator's efficiency can be strictly improved if we 
incorporate information about group membership.
Essentially, this captures a scenario in which a decision-maker
has the option of keeping all the information they currently have
available, and adding group membership on top.

As an interesting direction for possible further exploration, 
in this section we consider a related but distinct question:
if we have a limited budget of variables that we can measure
in our construction of an approximator,
is it ever worthwhile --- from an efficiency-maximizing perspective ---
to use group membership $\gp$ 
{\em instead of} a more informative variable $\x\^{i}$?
If we have a function $f$ that depends on $\x\^{i}$ but not on $\gp$,
such a situation would have the following striking implication: that in
optimally simplifying a function $f$ by consulting only a reduced
set of variables, we may end up with an incentive to use
group membership --- which is irrelevant to the actual value of $f$ ---
in place of another variable that actually affects the $f$-value.

As we will show next,
it is possible to construct examples of such functions.
While such a construction is related to the 
general result \rf{stmt:group-split} proved in Section \ref{sec:transform},
it addresses a distinct issue --- the preference for 
using group membership instead of other variables, 
rather than the efficiency benefits
of adding group membership to an existing set of variables.

Our construction in this section is essentially providing an existence result,
showing that such a phenomenon is possible.
It appears to be an interesting open question to characterize
more extensively
when this preference for group membership in the presence of a
budget constraint can arise, or to provide a robust set of
sufficient conditions for it to arise.

\xhdr{A construction}
For our construction, we will work with a natural 4-variable generalization
of the function from Figure \ref{fig:example} in 
Section \ref{sec:examples}, which we describe again here from
first principles for the sake of completeness.
For ease of exposition, we work out the example with a 
structured version of the function 
that does not satisfy the genericity condition
\rf{stmt:genericity}.
However, by subsequently perturbing the function values 
and probabilities by 
arbitrarily small amounts, we can obtain an example that
satisfies the genericity condition, and
for which the same conclusions hold.

We start with Boolean variables 
$\x\^{1}, \x\^{2}, \x\^{3}$ and a group membership variable $\gp$,
and we define $f$ to be the {\em majority function} 
on $\x\^{1}, \x\^{2}, \x\^{3}$, which takes the value
1 when a majority of the coordinates  $\x\^{1}, \x\^{2}, \x\^{3}$
are equal to 1.
That is, for a feature vector $\x = (\x\^{1}, \x\^{2}, \x\^{3})$, 
we define $f$ to take the value $1$ when 2 or 3 of the coordinates
$\x\^{i}$ are equal to 1, and to take the value 0 when
0 or 1 of the coordinates $\x\^{i}$ are equal to 1.

We assume (as in Section \ref{sec:examples}) that 
half the applicants belong to $A$ and half to $D$,
and the values of
the variables $\x\^{1}, \x\^{2}, \x\^{3}$ are set at random for each
individual in a way that reflects disadvantage: for a small $\eps > 0$,
and probability values
$p_1, p_2, p_3$ each equal to $1 - \eps$, 
an $A$-applicant has each $\x_i$ set equal
to 1 independently with probability $p_i$, and 
a $D$-applicant has each $\x_i$ set equal
to 1 independently with probability $q_i = 1 - p_i$.

As noted above, 
by perturbing the values of $f$ very slightly, and perturbing
the values of $p_1, p_2$, and $p_3$ very slightly as well, we can obtain
an example satisfying the genericity and disadvantage conditions,
and for which the subsequent arguments will also hold.
However, since the exposition is much cleaner with the
structured instance in which $f$ is precisely the majority function,
and all $p_i$ are equal to $1 - \eps$, we work out the consequences
of the example in this structured form.

\xhdr{Bounding the Number of Variables Used}
Now suppose we only cared about optimizing efficiency, not equity,
and for some constant $c$, we wanted to use an approximator that only 
consulted the values of $c$ of the variables.
Which approximator would be the best one to use?

If $c = 3$ (so that we are required to ignore one of the variables 
$\x\^{1}, \x\^{2}, \x\^{3}, \gp$), the answer is clear:
since $f$ is not affected by the value of $\gp$ once the values of
the other three variables are known, we can ignore $\gp$ and
still have a perfect approximation to $f$.

But what about at the other extreme, when $c = 1$?
Here we are choosing among four possible approximators:
$g_i$ (for $i = 1, 2, 3$) which only consults the value of $\x_i$;
and $g_0$, which only consults the value of $\gp$.
Since in this section
we are only concerned about efficiency, not equity, 
we will only ask --- for pairs of approximators $g$ and $h$ ---
whether one strictly improves the other in efficiency.

In the remainder of this section, we prove

\begin{stmt}
For the given function $f$, and a sufficiently small positive value of
the parameter $\eps$, the approximator 
$g_0$ strictly improves each of $g_1, g_2, g_3$ in efficiency.
\label{stmt:one-variable}
\end{stmt}

Before proceeding to the proof, let us note that \rf{stmt:one-variable}
captures the striking effect we were seeking from the construction.
Specifically, we have a function that depends 
on only three of its four variables
($x_1, x_2, x_3$ but not $\gp$).
Yet if we are told that we can only find out the value of one of
these four variables for a given individual, the optimal choice
is to select the ``irrelevant'' variable $\gp$ rather than any of the others.
This is because $\gp$ contains so much information about disadvantage ---
in the form of distributional information about the other variables ---
that it is more valuable for estimating $f$ than any one of the
variables that actually affect the value of $f$.
Thus, our construction here, like the general result of 
Section \ref{sec:transform}, 
shows how simplifying approximations to $f$ can have
the effect of transforming the underlying disadvantage into bias.

\prevs{stmt:one-variable}{
For each of $i = 1, 2, 3$, the approximator $g_i$ creates two cells:
$C_{i1}$, containing all applicants for whom $\x\^{i} = 1$, and
$C_{i2}$, containing all applicants for whom $\x\^{i} = 0$.
We have $\ms(C_{i1}) = p_i/2 + q_i/2 = 1/2$, 
and so $\ms(C_{i2}) = 1/2$ as well.
The approximator $g_0$ also creates two cells:
$C_{01}$, containing all applicants for whom $\gp = A$, and
$C_{02}$, containing all applicants for whom $\gp = D$.
Here too we have $\ms(C_{01}) = \ms(C_{02}) = 1/2$.

Thus, for all four approximators $g_i$ ($i = 0, 1, 2, 3$),
if we think of the average $f$-value of admitted applicant
$\efc_{g_i}(r)$ as a function of $r$, this function
maintains a constant value for all $r \leq 1/2$ as applicants from
the higher cell are admitted, and then
it decreases linearly to a shared value --- the average $f$-value
over the whole population --- at $r = 1$ as applicants from the
lower cell are admitted.
It follows that in order to show that 
$g_0$ strictly
improves each of $g_1, g_2, g_3$ in efficiency,
we only need to show that when we seek to admit precisely
$r = 1/2$ of the applicants, 
the average $f$-value admitted under $g_0$ is strictly
higher than under $g_1, g_2$, or $g_3$;
that is, $\efc_{g_0}(1/2) > \efc_{g_i}(1/2)$ for $i = 1, 2, 3$.

We thus turn to a comparison of $\efc_{g_0}(1/2)$ and $\efc_{g_i}(1/2)$
for $i = 1, 2, 3$.
For $i = 1, 2, 3$, the value $\efc_{g_i}(1/2)$ is the total
$f$-value of all applicants with $\x\^{i} = 1$, divided by
the normalizing constant $1/2$.  This is a sum of eight terms:
in the 16 rows of the look-up table that defines $f$, eight
of these rows have $\x\^{i} = 1$, and these are the rows that
contribute to the value $\efc_{g_i}(1/2)$.  That is, we have
$$\efc_{g_i}(1/2) = 2 \sum_{(\x,\gp): \x\^{i} = 1} \ms(\x,\gp) f(\x,\gp).$$
Since the sum has the same value for each of $g_1, g_2$, and $g_3$,
we evaluate it for $g_1$, using the following enumeration:
\begin{itemize}
\item Its largest term is $\ms(1,1,1,A) f(1,1,1,A) = (1 - \eps)^3$.
\item The next largest terms are
$\ms(1,1,0,A) f(1,1,0,A)$ and $\ms(1,0,1,A) f(1,0,1,A)$,
which are both equal to $\eps (1 - \eps)^2$.
\item The next largest terms after that are
$\ms(1,1,0,D) f(1,1,0,D)$ and $\ms(1,0,1,D) f(1,0,1,D)$,
which are both equal to $\eps^2 (1 - \eps)$.
\item The term $\ms(1,1,1,D) f(1,1,1,D)$ is equal to $\eps^3$.
\item The remaining two terms 
$\ms(1,0,0,A) f(1,0,0,A)$ and $\ms(1,0,0,D) f(1,0,0,D)$
are both equal to $0$.
\end{itemize}
Thus (recalling that there is also a factor of 2 in front of
the overall sum), we have
$$\efc_{g_i}(1/2) = 2 (1 - \eps)^3 + 4 \eps (1 - \eps)^2 + 4 \eps^2 (1 - \eps) + 2 \eps^3.$$

Now, for comparison, we evaluate $\efc_{g_0}(1/2)$,
which is the total $f$-value of all applicants with $\gp = A$, divided by
the normalizing constant $1/2$. 
This too is a sum of eight terms, as follows:
$$\efc_{g_0}(1/2) = 2 \sum_{(\x,\gp): \gp = A} \ms(\x,\gp) f(\x,\gp).$$
We can evaluate this sum as follows.
\begin{itemize}
\item Its largest term is $\ms(1,1,1,A) f(1,1,1,A) = (1 - \eps)^3$,
as in the previous case of $\efc_{g_i}(1/2)$.
\item It also contains the three terms 
$\ms(1,1,0,A) f(1,1,0,A)$, $\ms(1,0,1,A) f(1,0,1,A)$, and 
$\ms(0,1,1,A) f(0,1,1,A)$, each of which is equal to $\eps (1 - \eps)^2$.
\item The other four terms all have feature vectors $\x$ 
in which a majority of the 
coordinates $\x\^{i}$ are equal to 0;
therefore, $f$ evaluates to $0$ on these feature vectors, and
so each of these terms is 0.
\end{itemize}
Thus 
$$\efc_{g_0}(1/2) = 2 (1 - \eps)^3 + 6 \eps (1 - \eps)^2.$$

Comparing values by subtracting them, we have
$$\efc_{g_0}(1/2) - \efc_{g_i}(1/2) = 
2 \eps (1 - \eps)^2 - 4 \eps^2 (1 - \eps) - 2 \eps^3.$$
For sufficiently small $\eps > 0$, the first of these three terms
is arbitrarily larger than the other two, and hence the
difference is positive.
It follows that $\efc_{g_0}(1/2) > \efc_{g_i}(1/2)$.
As argued above, this is sufficient to show that
$g_0$ strictly improves $g_i$ in efficiency,
completing the proof of \rf{stmt:one-variable}.
}

\section{Further Related Work}

As discussed in Section \ref{sec:intro}, our work is connected
to the growing literatures on algorithmic fairness
\cite{barocas-big-data-disparate,corbett-davies-critical-review-fair,dwork-fairness-awareness,feldman-certifying-disparate}
and on interpretability
\cite{doshi-velez-science-interp,lipton-mythos-interp,zeng-ustun-interpretable}.
Within the literature on fairness, there has been a line of recent research
showing conflicts between different formal definitions of what it means
for a prediction function to be fair
\cite{berk-fairness-state-of-art,chouldechova-fair-prediction,corbett-davies-alg-decision-making,kleinberg-itcs17}; 
a key distinction between that work and
ours is that the tensions we are identifying arise from a syntactic
constraint on the form of the prediction function --- that it follow our
definition of simplicity --- rather than a fairness requirement on the 
output of the prediction function.
Kearns et al., in their research on {\em fairness gerrymandering} 
\cite{kearns-fairness-gerrymandering}, 
also consider the complexity of subsets evaluated by a classifier,
although they are not considering analogous formalizations of simplicity, and the goals of their work --- on auditing and learning-theoretic
guarantees in the presence of fairness properties --- lie in a
different direction from ours.
Finally, recent work has developed some of the equity benefits of explicitly taking group membership into account in ranking and classification rules \cite{corbett-davies-alg-decision-making,FryerLoury2013,KLMR-aea18, lipton-disparate},
although again without incorporating the simplicity properties of these rules in the analysis.

Our results also have connections with early economic models on
discrimination  (see \cite{fang2011theories} for a review and
references). Many of these models are based on
scenarios in which employers use race or some other protected attribute as a
statistical proxy for variables they do not observe (e.g.
\cite{arrow1973theory} and \cite{phelps1972statistical}). As in our
model, the disadvantaged group has a worse distribution of inputs; but
conditional on all inputs, the ground truth can be the same between
the advantaged and disadvantaged groups. A key issue in these models, however,
distinct from our work, is that
the decision-maker only observes a subset of inputs: Since these
unobserved variables are distributionally worse for the disadvantaged group,
membership in that group becomes a negative proxy, and employers
will discriminate against them in a statistical sense. 
This formalism can thus be viewed as a basic example of how omitting
variables from a model (in this case because they are unobserved) can lead to
discrimination.  
Our results, in the framework of simple models that
we define here, suggest that this link to discrimination is not 
specific to the problem of missing variables, 
but is inherent to the process of simplification much more generally. 
And through this more general approach, we see that the link
between simplicity and discrimination does not
even rely on the use of group membership as a proxy since, for example, 
our first main result applies even to simple functions that do not use
group membership as a variable.



\section{Conclusion}

Our main results articulate a tension between formal notions of
simplicity and equity, 
for functions used to rank individuals based on their features.
One of our key findings shows that
if such a function is structurally simple --- 
in a certain mathematical sense that
captures a number of current frameworks for building small or
interpretable models ---
then it can be replaced with a (more complex) function that
improves it both in performance and in equity.
In other words, the decision to use a simple rule should not necessarily be
viewed as a trade-off between performance and equity, but
as a step that necessarily sacrifices {\em both} properties
relative to other options in the design of a rule.
Our other main finding is that even when the true underlying function for ranking does not depend on an individual's membership in an advantaged or disadvantaged group, any non-trivial simplification of this function creates an incentive to nonetheless use this group membership information, and in a way that hurts the disadvantaged group.
These results point toward a further dimension in the 
connection between notions of fairness, simplicity, and interpretability,
suggesting an additional level of subtlety in the way they interact.

Our work suggests several further questions.
First, we have focused on a particular notion of simplicity;
while it is general enough to include a number of the main formalisms
used for constructing prediction algorithms, including variable
selection and decision trees (and it is motivated
in part by psychological notions of categories and conjunctive
concepts), it is clear that there are also other ways in which
we could try formalizing the notion of a simple model.
We view the set of potential approaches to formulating these questions as
quite broad and promising for exploration, and 
it would be interesting to understand the interaction of other
such definitions with notions of equity and fairness.

One common alternative formulation of simplicity is worth noting
in this respect:
{\em linear} approximators.
In particular, suppose we simplify $f(x)$, not by clustering
distinct inputs into cells, but by optimally approximating it using a
linear function $L(x)$. Linear approximation is not simplification in
the sense of our paper because $L$ can potentially take as many
distinct values as $f$ does. But $L$ does simplify in a different
sense: this potentially large set of distinct values is represented
compactly as a weighted sum of terms.  It is an interesting open
question whether our results could be extended to  model
simplification through linear approximation (or more generally
approximation with a restricted function class).  To appreciate one of
the challenges inherent in finding the right
formalism, note that linear approximations do not satisfy the
``truth-telling'' property of the approximators we consider:
for a value $y$ taken by a linear function $L$, 
if we look at the set of feature vectors $\x$
for which $L(\x) = y$, it is not the case in general 
that the average $f$-value in this set is $y$.
However, the values of a linear approximation satisfy other constrained structural properties, and
understanding how these interact with considerations
of equity is an interesting direction for further exploration.

Similarly, it would be natural to consider the effect of varying
other definitions in our framework;
for example, while the disadvantage condition we use is
motivated by a standard method for comparing distributions,
it would be interesting as noted earlier to understand what results follow
from alternate definitions of disadvantage.
Finally, our framework appears to have a rich mathematical
structure; for example, one could investigate
the space of approximators that are not strictly improvable 
as an object in itself, and to see what the resulting structure
suggests about the trade-offs we make when we choose to simplify
our models.

\subsection*{Acknowledgements}

  We thank Rediet Abebe, Solon Barocas, Fernando Delgado, Christian Hansen,
  Karen Levy, Jens Ludwig, Samir Passi, Manish Raghavan, Ashesh Rambachan,
  David Robinson, Joshua Schwartzstein, and Jann Spiess 
  for valuable discussions.
  The work has been supported in part by the MacArthur Foundation,
  the Sage Foundation, 
  a George C. Tiao Faculty Fellowship 
  at the University of Chicago Booth School, 
  and a Simons Investigator Award.


\begin{thebibliography}{10}

\bibitem{agan-ban-the-box}
Amanda Agan and Sonja Starr.
\newblock Ban the box, criminal records, and racial discrimination: A field
  experiment.
\newblock {\em Quarterly Journal of Economics}, 133(1):191--235, 2018.

\bibitem{arrow1973theory}
Kenneth Arrow.
\newblock The theory of discrimination.
\newblock {\em Discrimination in labor markets}, 3(10):3--33, 1973.

\bibitem{athey-monotone-likelihood}
Susan Athey.
\newblock Monotone comparative statics under uncertainty.
\newblock {\em Quarterly Journal of Economics}, 117(1):187--223, 2002.

\bibitem{barocas-big-data-disparate}
Solon Barocas and Andrew Selbst.
\newblock Big data's disparate impact.
\newblock {\em California Law Review}, 104, 2016.

\bibitem{berk-fairness-state-of-art}
Richard Berk, Hoda Heidari, Shahin Jabbari, Michael Kearns, and Aaron Roth.
\newblock Fairness in criminal justice risk assessments: {The} state of the
  art.
\newblock {\em Sociological Methods and Research}, 2018.

\bibitem{blyth-simpsons-paradox}
Colin Blyth.
\newblock On {Simpson's Paradox} and the sure-thing principle.
\newblock {\em Journal of the American Statistical Association}, 67(338), 1972.

\bibitem{chouldechova-fair-prediction}
Alexandra Chouldechova.
\newblock Fair prediction with disparate impact: {A} study of bias in
  recidivism prediction instruments.
\newblock {\em Big Data}, 2017.

\bibitem{chouldechova-child-hotline}
Alexandra Chouldechova, Diana~Benavides Prado, Oleksandr Fialko, and Rhema
  Vaithianathan.
\newblock A case study of algorithm-assisted decision making in child
  maltreatment hotline screening decisions.
\newblock In {\em Conference on Fairness, Accountability and Transparency,
  {FAT} 2018,}, pages 134--148, 2018.

\bibitem{corbett-davies-critical-review-fair}
Sam Corbett-Davies and Sharad Goel.
\newblock The measure and mismeasure of fairness: {A} critical review of fair
  machine learning.
\newblock Technical Report 1808.00023, arxiv.org, August 2018.

\bibitem{corbett-davies-alg-decision-making}
Sam Corbett-Davies, Emma Pierson, Avi Feller, Sharad Goel, and Aziz Huq.
\newblock Algorithmic decision making and the cost of fairness.
\newblock In {\em Proceedings of the 23rd Conference on Knowledge Discovery and
  Data Mining (KDD)}, 2017.

\bibitem{doshi-velez-science-interp}
Finale Doshi{-}Velez and Been Kim.
\newblock A roadmap for a rigorous science of interpretability.
\newblock Technical Report 1702.08608, arxiv.org, February 2017.

\bibitem{dwork-fairness-awareness}
Cynthia Dwork, Moritz Hardt, Toniann Pitassi, Omer Reingold, and Richard~S.
  Zemel.
\newblock Fairness through awareness.
\newblock In {\em Innovations in Theoretical Computer Science}, pages 214--226,
  2012.

\bibitem{fang2011theories}
Hanming Fang and Andrea Moro.
\newblock Theories of statistical discrimination and affirmative action: {A}
  survey.
\newblock In {\em Handbook of social economics}, volume~1, pages 133--200.
  Elsevier, 2011.

\bibitem{feldman-certifying-disparate}
Michael Feldman, Sorelle Friedler, John Moeller, Carlos Scheidegger, and Suresh
  Venkatasubramanian.
\newblock Certifying and removing disparate impact.
\newblock In {\em Proceedings of the 21st ACM SIGKDD International Conference
  on Knowledge Discovery and Data Mining}, KDD '15, pages 259--268, 2015.

\bibitem{FryerLoury2013}
Roland Fryer and Glenn Loury.
\newblock Valuing diversity.
\newblock {\em Journal of Political Economy}, 121(4):747--774, 2013.

\bibitem{garcia2001conjunctive}
Juan~A Garc{\'\i}a-Madruga, S~Moreno, N~Carriedo, F~Guti{\'e}rrez, and
  PN~Johnson-Laird.
\newblock Are conjunctive inferences easier than disjunctive inferences? {A}
  comparison of rules and models.
\newblock {\em The Quarterly Journal of Experimental Psychology: Section A},
  54(2):613--632, 2001.

\bibitem{goodman-gdpr}
Bryce Goodman and Seth~R. Flaxman.
\newblock European union regulations on algorithmic decision-making and a
  "right to explanation".
\newblock {\em {AI} Magazine}, 38(3):50--57, 2017.

\bibitem{greenwald-implicit-social-cognition}
Anthony~G. Greenwald and Mahzarin~R. Banaji.
\newblock Implicit social cognition: {Attitudes}, self-esteem, and stereotypes.
\newblock {\em Psychological Review}, 102(1):4--27, 1995.

\bibitem{hastie-stat-learning-book}
Trevor Hastie, Robert Tibshirani, and Jerome Friedman.
\newblock {\em The Elements of Statistical Learning: {Data} Mining, Inference,
  and Prediction}.
\newblock Springer, 2009.

\bibitem{hopkins-ratio-orderings}
Ed~Hopkins and Tatiana Kornienko.
\newblock Ratio orderings and comparative statics.
\newblock Technical Report~91, Edinburgh School of Economics Discussion Paper
  Series, 2003.

\bibitem{kearns-fairness-gerrymandering}
Michael~J. Kearns, Seth Neel, Aaron Roth, and Zhiwei~Steven Wu.
\newblock Preventing fairness gerrymandering: {Auditing} and learning for
  subgroup fairness.
\newblock In {\em Proceedings of the 35th International Conference on Machine
  Learning, {ICML}}, pages 2569--2577, 2018.

\bibitem{kleinberg-bail-qje}
Jon Kleinberg, Hima Lakkaraju, Jure Leskovec, Jens Ludwig, and Sendhil
  Mullainathan.
\newblock Human decisions and machine predictions.
\newblock {\em Quarterly Journal of Economics}, 133(1):237--293, 2018.

\bibitem{kleinberg2015prediction}
Jon Kleinberg, Jens Ludwig, Sendhil Mullainathan, and Ziad Obermeyer.
\newblock Prediction policy problems.
\newblock {\em American Economic Review}, 105(5):491--95, 2015.

\bibitem{KLMR-welfare-function}
Jon Kleinberg, Jens Ludwig, Sendhil Mullainathan, and Ashesh Rambachan.
\newblock Algorithmic bias and the social welfare function: {Regulating}
  outputs versus regulating algorithms, 2018.
\newblock Working paper.

\bibitem{KLMR-aea18}
Jon Kleinberg, Jens Ludwig, Sendhil Mullainathan, and Ashesh Rambachan.
\newblock Algorithmic fairness.
\newblock {\em AEA Papers and Proceedings}, pages 22--27, 2018.

\bibitem{kleinberg-itcs17}
Jon Kleinberg, Sendhil Mullainathan, and Manish Raghavan.
\newblock Inherent trade-offs in the fair determination of risk scores.
\newblock In {\em 8th Innovations in Theoretical Computer Science Conference,
  {ITCS}}, 2017.

\bibitem{leyens-stereotypes-book}
Jacques-Philippe Leyens, Vincent~Y.A. Yzerbyt, and Georges Schadron.
\newblock {\em Stereotypes and Social Cognition}.
\newblock Sage Publications, 1994.

\bibitem{lipton-mythos-interp}
Zachary~C. Lipton.
\newblock The mythos of model interpretability.
\newblock {\em Communications of the ACM}, 61(10):36--43, 2018.

\bibitem{lipton-disparate}
Zachary~Chase Lipton, Alexandra Chouldechova, and Julian McAuley.
\newblock Does mitigating {ML}'s impact disparity require treatment disparity?
\newblock Technical Report 1711.07076, arxiv.org, November 2017.

\bibitem{milgrom-good-news}
Paul~R. Milgrom.
\newblock Good news and bad news: {R}epresentation theorems and applications.
\newblock {\em Bell Journal of Economics}, 12:380--391, 1981.

\bibitem{mullainathan-categories}
Sendhil Mullainathan.
\newblock Thinking through categories, 2000.
\newblock Working paper.

\bibitem{mullainathan-coarse-thinking}
Sendhil Mullainathan, Joshua Schwartzstein, and Andrei Shleifer.
\newblock Coarse thinking and persuasion.
\newblock {\em Quarterly Journal of Economics}, 123(2):577--619, 2008.

\bibitem{phelps1972statistical}
Edmund~S Phelps.
\newblock The statistical theory of racism and sexism.
\newblock {\em American Economic Review}, pages 659--661, 1972.

\bibitem{rockoff-teacher-recruiting}
Jonah~E. Rockoff, Brian~A. Jacob, Thomas~J. Kane, and Douglas~O. Staiger.
\newblock Can you recognize an effective teacher when you recruit one?
\newblock {\em Education Finance and Policy}, 6(1):43--74, 2011.

\bibitem{rosch1978cognition}
Eleanor Rosch and Barbara~Bloom Lloyd.
\newblock {\em Cognition and categorization}.
\newblock Lawrence Erlbaum Associates Hillsdale, NJ, 1978.

\bibitem{stevenson-assessing-risk-assessment}
Megan Stevenson.
\newblock Assessing risk assessment in action.
\newblock {\em Minnesota Law Review}, 103, 2018.

\bibitem{wolfstetter-microeconomics-book}
Elmar Wolfstetter.
\newblock {\em Topics in Microeconomics}.
\newblock Cambridge University Press, 1999.

\bibitem{zeng-ustun-interpretable}
Jiaming Zeng, Berk Ustun, and Cynthia Rudin.
\newblock Interpretable classification models for recidivism prediction.
\newblock {\em Journal of the Royal Statistical Society: Series A},
  180(3):689--722, 2017.

\end{thebibliography}


\bigskip

\section*{Appendix A. Every Approximator Can Be Improved by a Maximal Approximator}

In this appendix, we provide a proof of \rf{stmt:maximal-improve},
that for every $f$-approximator, there is a maximal $f$-approximator
that weakly improves it.
Thus, we start with an arbitrary $f$-approximator $g$, consisting
of cells $C_1, \ldots, C_d$, with $d \leq B$ for our absolute bound
on the number of allowable cells.  Each cell $C_i$ is described
by a vector $\phi_i = (\phi_i(\xb_1), \ldots, \phi_i(\xb_m))$
where $\xb_1, \ldots, \xb_m$ is an enumeration of all $m = 2^{k+1}$ rows,
and $\phi_i(\xb_j)$ specifies the measure of row $\xb_i$ assigned to
cell $C_i$.

To find a maximal $f$-approximator that weakly improves $g$,
we will work with a representation of $f$-approximators
as points in Euclidean space, so that we can eventually use an 
argument based on compactness and continuity.
We say that an {\em $f$-synthesizer} is a vector of values
$\psi = (\psi_{ij} : 1 \leq i \leq B; 1 \leq j \leq m)$, where
$\psi_{ij}$ is intended to represent the value
$\phi_i(\xb_j)$ associated with $g$.
(Below, we will deal with the issue that $\psi$ is indexed all the
way out to $B$, while $g$ may have only $d < B$ cells.)
For $\psi$ to faithfully represent the values $\phi_i(\xb_j)$,
we impose the following constraints on it.
\begin{itemize}
\item $\psi_{ij} \geq 0$ for all $1 \leq i \leq B$ and $1 \leq j \leq m$.
\item $\sum_{i=1}^B \psi_{ij} = \ms(\xb_j)$ for all $1 \leq j \leq m$,
so that each row is completely allocated across the cells.
\item Finally, the cells $C_1, \ldots, C_d$ of $g$ are 
sorted in descending order of
$\gfn(C_i) = { \sum_{\xb} \phi_i(\xb) f(\xb) } / { \sum_{\xb} \phi_i(\xb) }$;
the condition $\gfn(C_h) \geq \gfn(C_i)$ for $h \leq i$ can be equivalently
written as
\begin{equation*}
\left({ \sum_{\xb} \phi_h(\xb) f(\xb) }\right) \left({ \sum_{\xb} \phi_i(\xb) }\right)
-
\left({ \sum_{\xb} \phi_i(\xb) f(\xb) }\right) \left({ \sum_{\xb} \phi_h(\xb) }\right) \geq 0.
\end{equation*}
We therefore impose the following constraint on $\psi$, for all
$1 \leq i \leq j \leq B$.
\begin{equation}
\left({ \sum_{j=1}^m \psi_{hj} f(\xb_j) }\right) \left({ \sum_{j=1}^m \psi_{ij} }\right)
-
\left({ \sum_{j=1}^m \psi_{ij} f(\xb_j) }\right) \left({ \sum_{j=1}^m \psi_{hj} }\right)
\geq 0.
\label{eq:cell-order}
\end{equation}
As noted above, this naturally represents approximators that
have exactly $B$ cells.
For approximators that have $d < B$ cells, we adopt a slightly
unusual convention that makes the representation in Euclidean space easier.
In particular, if $g$ has $d < B$ cells, then we also declare that $g$
has $B - d$ {\em empty cells}.
Each empty cell $C_i$ has associated vector
$\phi_i = 0$, and it can come anywhere in the sorted order.
We will not attempt to define a value $\gfn(C_i)$ for an empty cell;
but this will not pose a problem, since no portion of the population
belongs to this cell.
Now, wherever we place the empty cells in the sorted order, they
will satisfy Inequality (\ref{eq:cell-order}) (since they will produce
a left-hand side of $0$ with respect to any other cell).
\end{itemize}

The intersection of these constraints defines the set of
$f$-synthesizers $K \subseteq \R^{mB}$.
Note that the set $K$ is a closed and bounded subset
of Euclidean space, and hence compact.
Every $f$-approximator $g$ with $d$ cells can be naturally mapped to an
$f$-synthesizer in $K$: we simply concatenate $B - d$ empty cells
to the end of $g$'s list of cells, and write $\psi_{ij}$ for $\phi_i(\xb_j)$.
We can check that all the constraints are satisfied.
Conversely, given any $f$-synthesizer $\psi$, we can create
an $f$-approximator $g$ as follows:
for every $i$ such that $\sum_{j=1}^m \psi_{ij} > 0$, we create
a cell of $g$ with $\phi_i(\xb_j) = \psi_{ij}$.
These cells will be arranged in decreasing order of
$\gfn$-value, and every row will be completely allocated
across the cells.


For a vector $\psi \in K$, let $g(\psi)$ be the approximator
produced by this construction, and let $\lambda(\psi)$ be the 
univariate function $\efd_{g(\psi)}(\cdot)$.
As discussed earlier in the text, 
this function $\efd_{g(\psi)}(\cdot) = \lambda(\psi)$
is piecewise constant, with an interval
over which it is constant for each cell, and a finite set
of points of discontinuity corresponding to the points between
consecutive cells.
Let $\Lambda_r(\psi)$ be the value of $\efc_g(r)$ for this 
$f$-approximator $g(\psi)$.
If $\psi\^1, \psi\^2, \psi\^3, \ldots$ is a convergent sequence in $K$
with limit $\psi^*$, then the functions
$\lambda(\psi\^1), \lambda(\psi\^2), \lambda(\psi\^3), \ldots$ 
converge pointwise to the
function $\lambda(\psi^*)$ except possibly at its finite set
of points of discontinuity.
It follows that the values 
$\Lambda_r(\psi\^1), \Lambda_r(\psi\^2), \Lambda_r(\psi\^3), \ldots$
converge to $\Lambda_r(\psi^*)$.

We conclude two things from this argument.
First, the function $\Lambda_r(\cdot)$ is a continuous function on $K$,
and second, for any $\basepsi \in K$, the set $L(r,\basepsi)$ of all $\psi$ 
for which $\Lambda_r(\psi) \geq \Lambda_r(\basepsi)$ is
a closed subset of $K$.
Moreover, if we define $\Gamma_r(\psi)$ to be 
the value of $\eqc_{g(\psi)}(r)$, then
the same argument can be applied to $\Gamma_r$, showing that
$\Gamma_r(\cdot)$ is continuous, and the set
$M(r,\basepsi)$ of all $\psi$ 
for which $\Gamma_r(\psi) \geq \Gamma_r(\basepsi)$ 
is a closed subset of $K$.

Now, given an $f$-approximator $g$, 
we would like to use these definitions to construct
a maximal $f$-approximator that weakly improves $g$.
First, we choose an $f$-synthesizer $\basepsi$ such that
$g(\basepsi) = g$.
Next, we define a set intended to represent all $f$-approximators
that weakly improve on $g(\basepsi)$.
Specifically, we define
$$N(\basepsi) = K \cap \bigcap_{0 < r < 1} L(r,\basepsi) \cap
\bigcap_{0 < r < 1} M(r,\basepsi).$$
This is an intersection of closed sets, and hence it is closed;
since it also bounded, it is a compact set.
It also non-empty, since it contains $\basepsi$.

Finally, for $\psi \in K$, let
$\Omega(\psi) = \int_0^1 \efc_{g(\psi)}(t) ~ dt$.
This is a continuous function of $\psi$;
therefore, since $N(\basepsi)$ is a compact set,
the maximum value of $\Omega$ over the set $N(\basepsi)$
is assumed at some non-empty subset of $N(\basepsi)$.
Let $\psi^+$ be a point in this subset.

Consider the $f$-approximator $g^+ = g(\psi^+)$;
we claim that $g^+$ is maximal.
For if not, there would be a point $\psi' \in N(\basepsi)$ such that
$\efc_{g(\psi')}(r) \geq \efc_{g(\psi^+)}(r)$ for all $r$, and
$\efc_{g(\psi')}(r^*) > \efc_{g(\psi^+)}(r^*)$ for some $r^*$.
Since $\efc_{g(\psi')}(\cdot)$ and $\efc_{g(\psi^+)}(\cdot)$
are continuous functions, it would follow that
$\Omega(\psi') > \Omega(\psi^+)$, 
contradicting the assumption that $\Omega$
assumes its maximum value in $N(\basepsi)$ at the point $\psi^+$.

Since we have constructed a maximal $f$-approximator $g^+$
that weakly improves $g$, this completes the proof of
\rf{stmt:maximal-improve}

\section*{Appendix B. Comparing Random Variables}

In this section, we provide a proof of \rf{stmt:rv-compare-exp}.
It is useful to state it in a more expansive form that 
brings in an additional property.
The resulting formal statement is standard in the literature
on comparing random variables \cite{hopkins-ratio-orderings,wolfstetter-microeconomics-book},
and our proof is purely for the sake of completeness, to 
cast it in our current discrete formalism.

\vskip \belowdisplayskip

{\em (B.1) (See e.g. \cite{hopkins-ratio-orderings,wolfstetter-microeconomics-book})
Consider two discrete random variables $\rvd$ and $\rva$,
each of which takes values in $\{\rvval_1, \rvval_2, \ldots, \rvval_n\}$,
with $\rvval_1 < \rvval_2 < \cdots < \rvval_n$ and $n > 1$.
Let $\rvdprob_i = \Prb{\rvd_i = \rvval_i}$ and $\rvaprob_i = \Prb{\rva_i = \rvval_i}$;
so $\sum_{i = 1}^n \rvdprob_i = \sum_{i = 1}^n \rvaprob_i = 1$, and
$\Exp{\rvd} = \sum_{i = 1}^n \rvdprob_i \rvval_i$ and
$\Exp{\rva} = \sum_{i = 1}^n \rvaprob_i \rvval_i$.
We will assume that $\rvdprob_i > 0$ and $\rvaprob_i > 0$ for all $i$.

Consider three different comparisons between $\rva$ and $\rvd$:
\begin{itemize}
\item[(i)] {\em Expectation Dominance}: $\Exp{\rva} > \Exp{\rvd}$.
\item[(ii)] {\em First-Order Stochastic Dominance}:
For all $t$ such that $\rvval_1 \leq t < \rvval_n$, we have
$\Prb{\rva > t} > \Prb{\rvd > t}$.
\item[(iii)] {\em Likelihood Ratio Dominance}:
The sequence of ratios $\{\rvaprob_i/\rvdprob_i\}$ is strictly monotonically increasing.
\end{itemize}

For all pairs of random variables $\rva$ and $\rvd$ as above,
condition (iii) implies condition (ii), and condition (ii) implies condition (i).
}

\vskip 0.1in \noindent {\em Proof of (B.1).} 
We define
$\cdd_i = \Prb{\rvd \leq \rvval_i} = \sum_{\ell=1}^i \rvdprob_{\ell}$ and
$\cda_i = \Prb{\rva \leq \rvval_i} = \sum_{\ell=1}^i \rvaprob_{\ell}$; note that then
$1 - \cdd_i = \Prb{\rvd > \rvval_i} = \sum_{\ell=i+1}^n \rvdprob_{\ell}$ and
$1 - \cda_i = \Prb{\rva > \rvval_i} = \sum_{\ell=i+1}^n \rvaprob_{\ell}$.

We first show that (iii) implies condition (ii).
Since $\sum_{i=1}^n \rvdprob_i = \sum_{i=1}^n \rvaprob_i = 1$, we cannot have
$\rvdprob_i \geq \rvaprob_i$ for all $i$ or $\rvdprob_i \leq \rvaprob_i$ for all $i$.
Thus, by Likelihood Ratio Dominance, we have $\rvdprob_i > \rvaprob_i$ up to some $i = i^*$,
and then $\rvdprob_i \leq \rvaprob_i$ for $i > i^*$.
Let $\eps = \sum_{i = 1}^{i^*} (\rvdprob_i - \rvaprob_i)$.
Note that since $\sum_{i=1}^n \rvdprob_i = \sum_{i=1}^n \rvaprob_i = 1$,
we also have $\eps = \sum_{i = i^* + 1}^{n} (\rvaprob_i - \rvdprob_i)$, and
hence $\sum_{i = i^* + 1}^{\ell} (\rvaprob_i - \rvdprob_i) < \eps$
for $\ell < n$.

We would like to show that $\cda_i < \cdd_i$ for all $i < n$.
For $i \leq i^*$, this follows simply because $\rvaprob_i < \rvdprob_i$ for
all such $i$.
For $i > i^*$, we have
$$\sum_{j=1}^i (\rvaprob_i - \rvdprob_i)
  = \sum_{j=1}^{i^*} (\rvaprob_i - \rvdprob_i) + \sum_{j=i^* + 1}^{i} (\rvaprob_i - \rvdprob_i)
  < -\eps + \eps = 0,$$
and hence $\cda_i < \cdd_i$ for $i > i^*$ as well.
This shows that condition (iii) implies condition (ii).

We now show that condition (ii) implies condition (i).
Let $\eps_i = \rvval_i - \rvval_{i-1} > 0$.
We have
\begin{eqnarray*}
\Exp{\rvd} & = & \sum_{i = 1}^n \rvdprob_i \rvval_i \\
& = & \rvdprob_1 \rvval_1 + \rvdprob_2 (\rvval_1 + \eps_2) + \rvdprob_3 (\rvval_1 + \eps_2 + \eps_3)
+ \cdots  \\
& & ~~~ + ~ \rvdprob_n(\rvval_1 + \eps_2 + \eps_3 + \cdots + \eps_n) \\
& = & \rvval_1 + (1 - \cdd_1) \eps_2 + (1 - \cdd_2) \eps_3 + \cdots + (1 - \cdd_{n-1}) \eps_n,
\end{eqnarray*}
where we pass from the first line to the second line by
writing $\rvval_i$ as $\rvval_1 + \sum_{j=2}^i \eps_j$, and we pass
from the second line to the third line by collecting together all
the $\rvdprob_i$ that are multiplied by each $\eps_j$.

Analogously, we have
$$\Exp{\rva} = \rvval_1 + (1 - \cda_1) \eps_2 + (1 - \cda_2) \eps_3 + \cdots + (1 - \cda_{n-1}) \eps_n.$$
Now, using the fact that $n > 1$ and $1 - \cda_i > 1 - \cdd_i$ for all $i < n$,
we obtain
$\Exp{\rva} > \Exp{\rvd}.$
\rule{2mm}{2mm}

\end{document}